\documentclass{article}

\usepackage[preprint]{neurips_2025}

\usepackage[utf8]{inputenc} % allow utf-8 input
\usepackage[T1]{fontenc}    % use 8-bit T1 fonts
\usepackage{hyperref}       % hyperlinks
\usepackage{url}            % simple URL typesetting
\usepackage{booktabs}       % professional-quality tables
\usepackage{amsfonts}       % blackboard math symbols
\usepackage{nicefrac}       % compact symbols for 1/2, etc.
\usepackage{microtype}      % microtypography
\usepackage{xcolor}         % colors
\usepackage{graphicx}
\usepackage{multirow}
\usepackage{amsmath}
\usepackage{colortbl}
\usepackage{cleveref}
\usepackage[ruled,vlined]{algorithm2e}

\usepackage[export]{adjustbox}
\usepackage{enumitem}
\usepackage{xcolor}

\usepackage{longtable}
\usepackage{tabularx}
\usepackage{booktabs}
\usepackage{amssymb} % For \checkmark
\usepackage{pifont}  % For \ding
\usepackage[dvipsnames]{xcolor}  % For colored symbols

\newcommand{\cmark}{\ding{51}} % Define custom command
\newcommand{\xmark}{\ding{55}} % Define custom command

\title{Semantic Context-aware Modality Fusion Transformer (SCOUT): A Context-Aware Multimodal Transformer for Concept-Grounded Pathology Report Generation}
\newcommand{\sbu}{\textsuperscript{1}}

\author{%
  Suryakant Singh\sbu
  \And
  Saarthak Kapse\sbu\And
  Joel Saltz\sbu \And
  Prateek Prasanna\sbu
}

\date{%
  \begin{center}
    \textbf{\sbu Stony Brook University}
  \end{center}
}

\begin{document}

\maketitle

\begin{abstract}

Whole-slide pathology report generation requires models to integrate localized histomorphology, global tissue context, and diagnostically relevant semantic information, yet existing approaches typically rely on fixed pretrained visual representations and may fail to represent key diagnostic concepts. Here we present \textbf{SCOUT}: \textbf{S}emantic \textbf{C}ontext-aware m\textbf{O}dality f\textbf{U}sion \textbf{T}ransformer, a concept-grounded multimodal framework that integrates local histological patterns, whole-slide context, and expert-curated diagnostic concepts. SCOUT maintains an evolving visual representation together with recursively updated slide- and concept-conditioned representations, allowing visual and contextual information to iteratively co-evolve across encoder depth. During decoding, separate attention pathways attend to these complementary streams before an adaptive multimodal fusion module combines them for each generated token.
 
We evaluated SCOUT on TCGA-BRCA, HistAI, and REG-2025, comprising more than 20,900 WSI-report pairs across more than seven cancer types. Using common CONCHv1.5 features and evaluation protocols, SCOUT outperformed WSI-Caption, HistGen, and Bi-Gen, improving BLEU scores by 9.6\% on average, METEOR by 11.0\%, and ROUGE-L by 3.6\% relative to the strongest competing method. On REG-2025, SCOUT additionally improved the Clinical Report Quality Score by 5.6\%, indicating better preservation of clinically relevant report attributes beyond lexical similarity. Ablations showed complementary contributions from iterative context refinement and adaptive fusion, while learned gates provided inspectable token-level estimates of modality use.

Our results suggest that progressive contextual conditioning is effective across heterogeneous pathology report generation settings and provides a flexible framework for integrating domain knowledge into vision-language models for computational pathology.

% 175 word version

% Whole-slide pathology report generation requires integration of localized histomorphology, global tissue context, and diagnostically relevant semantic information, yet existing approaches often rely on fixed pretrained visual representations which may fail to represent key diagnostic concepts. Here we present \textbf{SCOUT}: \textbf{S}emantic \textbf{C}ontext-aware m\textbf{O}dality f\textbf{U}sion \textbf{T}ransformer, a concept-grounded multimodal framework that combines local histological patterns, whole-slide context, and expert-curated diagnostic concepts. SCOUT maintains an evolving visual representation together with recursively updated slide- and concept-conditioned streams, enabling iterative co-evolution across encoder depth. During decoding, separate attention pathways preserve these complementary streams before adaptive multimodal fusion combines them for each generated token. 

% We evaluated SCOUT on TCGA-BRCA, HistAI, and REG-2025, comprising more than 20,900 WSI-report pairs across more than seven cancer types. SCOUT outperformed WSI-Caption, HistGen, and Bi-Gen, improving BLEU by 9.6\% on average, METEOR by 11.0\%, and ROUGE-L by 3.6\%. On REG-2025, SCOUT additionally improved Clinical Report Quality Score by 5.6\%. These results indicate that progressive contextual conditioning and adaptive multimodal fusion are effective strategies for pathology report generation.

\end{abstract}

%%%% Main Text %%%%%%%%%%%%%%%%%%%%%%%%%%%%%%%%%%%%%%%%%%%%%

\section{Introduction}

\begin{figure}[t]
    \centering
    \includegraphics[width=1\linewidth]{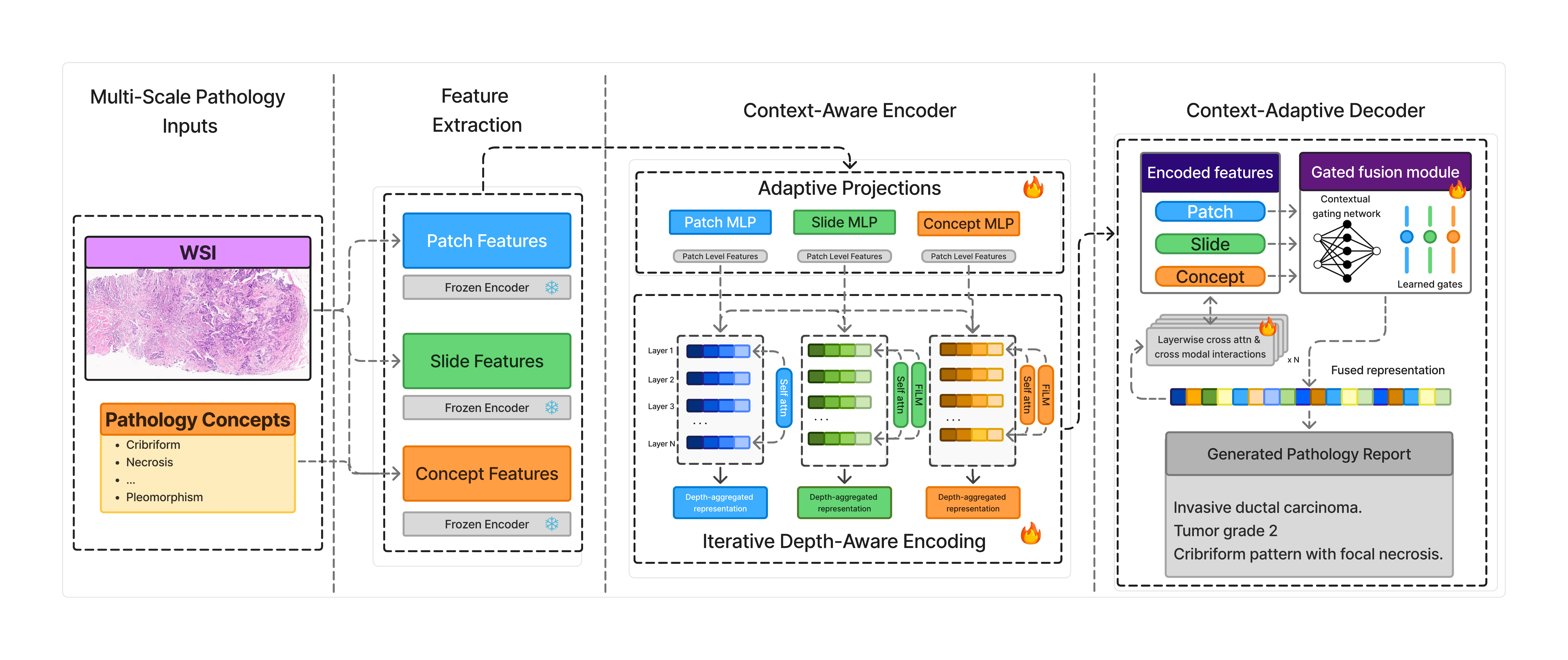}
    \caption{\textbf{Overview of SCOUT for pathology report generation.}
    Patch-level visual features are extracted using frozen CONCHv1.5 features,
    while TITAN and GECKO provide complementary slide-level and concept-level
    representations. Modality-specific adapters map these heterogeneous features
    to a shared latent space. Within the context-aware encoder, an evolving patch backbone is used to construct parallel slide-conditioned and
    concept-conditioned feature streams. The corresponding contextual states are
    recursively updated across encoder depth, and each of the three feature
    streams is independently aggregated using learned layer weights. The decoder
    then attends separately to the patch, slide-conditioned, and
    concept-conditioned representations and combines them using feature-wise
    adaptive gated fusion during autoregressive report generation.}
    \label{fig:architecture_overview}
\end{figure}

% Automated generation of structured reports from medical images has emerged as a promising application of multimodal machine learning, with potential to reduce documentation burden and improve consistency in clinical workflows\cite{histgen,wsi_caption,bigen, lucassen2025pathology, hu2025pathology,jin2026grounded,gao2025s2d}. In computational pathology, this task is particularly challenging\cite{gamper2021multiple,huang2023visual,lu2023visual}: pathology reports synthesize fine-grained cellular observations, global tissue architecture, and standardized diagnostic concepts into coherent narratives that are both descriptive and explanatory. Report generation in this setting therefore requires not only visual understanding, but also the integration of multi-scale evidence with structured medical knowledge.

Automated generation of structured reports from medical images has emerged as a promising application of multimodal machine learning, with potential to reduce documentation burden, improve reporting consistency, and support large-scale analysis of clinical imaging data \cite{histgen,wsi_caption,bigen,histogpt,lucassen2025pathology,hu2025pathology,jin2026grounded,gao2025s2d}. In computational pathology, this task is particularly challenging because whole-slide images (WSIs) contain diagnostically relevant information across markedly different spatial scales. Pathology reports must integrate local cellular morphology, global tissue architecture, and standardized diagnostic concepts into concise narratives that are both descriptive and clinically meaningful\cite{gamper2021multiple,huang2023visual,lu2023visual}.

Recent advances in multimodal learning and pathology vision-language models have enabled increasingly fluent report generation from whole-slide images (WSIs). However, most existing approaches rely on fixed patch level visual representations extracted prior to language modeling\cite{wsi_caption,histogpt,gamper2021multiple,huang2023visual,lu2023visual} or incorporate additional modalities derived from the same underlying patch features, without any contextual adaptation of visual representations. Such designs implicitly assume that these visual representations are universally optimal for all downstream linguistic and reasoning tasks. This assumption is particularly limiting in pathology, where diagnostically relevant features vary substantially across disease types, magnification levels, and reporting conventions.

A central limitation of current pathology report generation methods is the limited contextual adaptation of pretrained visual representations during downstream report generation. Local patch-level features often capture fine cellular detail but lack global context, while slide-level representations summarize tissue architecture at the cost of spatial specificity. This design separates visual representation learning from the contextual and semantic information needed to interpret the image. Such separation is restrictive in pathology, where the significance of a local pattern depends strongly on its surrounding tissue architecture, disease context, and relationship to established diagnostic concepts. Models may generate fluent reports while failing to accurately ground statements in relevant visual evidence, and often provide limited insight into how diagnostic concepts influence predictions. In contrast, human pathology practice involves iterative reasoning, where visual impressions are continuously refined based on global context and domain knowledge. Capturing this process computationally requires models in which visual representations are dynamically modulated by contextual and semantic signals throughout the learning pipeline.

Existing methods have begun to address parts of this problem. Hierarchical approaches such as HistGen combine local and global visual information, whereas knowledge-enhanced methods such as Bi-Gen incorporate retrieved textual context \cite{histgen,bigen}. These strategies improve feature aggregation and semantic richness, but contextual information is generally introduced after visual features have already been formed or is incorporated as an auxiliary source during fusion. Consequently, they do not directly model how global tissue context and diagnostic concepts can progressively influence the interpretation of local morphology.

Here we introduce \textbf{SCOUT} (\textbf{S}emantic \textbf{C}ontext-aware m\textbf{O}dality f\textbf{U}sion \textbf{T}ransformer), a multimodal framework that jointly models patch-level morphology, whole-slide context, and explicit histopathologic diagnostic concepts. 
The central idea is to maintain local visual, global contextual, and concept-conditioned representations as distinct but interacting information streams throughout the encoder. At successive encoder layers, the co-evolving patch representations are used to construct slide-conditioned and concept-conditioned feature streams. These contextual streams are pooled to
update the corresponding slide and concept states, which subsequently condition the next encoder depth. This iterative process allows contextual representations to co-evolve together with increasingly transformed visual features rather than remaining fixed auxiliary inputs. Learned depth-wise aggregation then independently combines representations from different encoder layers for each stream.

During report generation, SCOUT maintains separate attention pathways for patch-, slide-, and concept-level representations.  A feature-wise multi-head gating mechanism adaptively combines the resulting cross-attended representations for each generated token. This enables the model to control how strongly each information source contributes to each generated token, allowing the model to vary its reliance on fine-grained morphology, global tissue organization, and diagnostic concepts across different parts of the report. This architecture therefore separates two complementary forms of multimodal interaction: iterative contextual updating during encoding and adaptive representation fusion during decoding. The learned modality weights provide token-level estimates of how visual and semantic information contribute during report generation. By jointly incorporating local visual evidence, global contextual information, and structured diagnostic concepts, the framework is designed to improve preservation of clinically relevant report content beyond lexical fluency alone. 

We evaluate SCOUT across three heterogeneous pathology report-generation
datasets and compare it with recent WSI report-generation approaches using
fixed data partitions, common patch representations, and standardized
preprocessing. In addition to conventional language-generation metrics, we
evaluate structured clinical report fidelity using the Clinical Report Quality
Score (CRQS). Ablation experiments examine the contributions of contextual
representation updating and adaptive fusion, while cohort-level gate analysis
and case-level cross-attention visualizations characterize how the three
representation streams are used during generation. Table ~\ref{tab:method_comparison} summarizes the main architectural distinctions between SCOUT and representative pathology report generation methods and Fig.~\ref{fig:architecture_overview} provides an overview of the model. Detailed definitions of the encoder, decoder, feature-extraction pipeline, and training procedure are provided in the Methods and Appendix.

\begin{table}[t]
\centering

\caption{\textbf{Comparison of representative pathology report generation
architectures.} SCOUT differs from prior approaches by maintaining explicit
patch, slide-conditioned, and concept-conditioned representation streams,
recursively updating contextual states across encoder depth, and adaptively
combining the streams during token generation.}
\small
\setlength{\tabcolsep}{4pt}
\renewcommand{\arraystretch}{1}
\begin{tabular}{lccccc}
\toprule
\textbf{Method} & \textbf{Multi-scale Visual input} & \textbf{Concept Use} & \textbf{Contextual Feature Refinement} & \textbf{Adaptive token-level fusion} \\
\midrule
WSI-Caption & \textcolor{red}\xmark (patch) & \textcolor{red}\xmark (None) & None & \textcolor{red}\xmark \\
HistGen & \textcolor{ForestGreen}\cmark (hierarchical) & \textcolor{red}\xmark (Implicit) & Encoder (Aggregation) & \textcolor{red}\xmark \\
Bi-Gen & \textcolor{red}\xmark (patch + patch) & \textcolor{ForestGreen}\cmark (Retrieval-based) & Encoder & \textcolor{red}\xmark \\
\textbf{SCOUT} & \textbf{\textcolor{ForestGreen}\cmark (patch + slide)} & \textbf{\textcolor{ForestGreen}\cmark (Explicit)} & \textbf{Encoder + Decoder} & \textbf{\textcolor{ForestGreen}\cmark} \\
\bottomrule
\end{tabular}
\label{tab:method_comparison}
\end{table}

Together, the combination of multi-scale pathology representations, iterative parallel co-evolution of visual and contextual features, and adaptive multimodal fusion provides a general framework for integrating heterogeneous sources of information during conditional text generation. By maintaining complementary representation streams throughout encoding and dynamically allocating their contributions during decoding, SCOUT separates contextual representation refinement from downstream multimodal fusion. Although we develop and evaluate the framework for whole-slide pathology report generation, its modular design may also be applicable to other medical image-to-text tasks in which information from multiple visual scales and structured semantic sources must be integrated to generate clinically meaningful narratives.

%All code required to reproduce the experiments in this study is available at our GitHub repository: \url{https://github.com/surykntsingh/SCOUT}.

\section{Results}

\subsection{Performance across three heterogeneous datasets}

% We evaluate the proposed framework on three benchmark datasets spanning heterogeneous pathology domains and reporting styles. Tables ~\ref{tab:brca}, ~\ref{tab:reg} and ~\ref{tab:histai} summarize the performance compared to state-of-the-art report generation methods, including MI-Gen, HistGen, and BiGen, using identical CONCHv1.5-derived visual features as input. Across all datasets, our method consistently outperforms prior approaches on standard language generation metrics. Notably, we observe improvements across BLEU-1 to BLEU-4, METEOR, and ROUGE-L, indicating gains in both lexical fidelity and semantic coherence. 

\begin{table*}[t]
\centering

\caption{\textbf{Characteristics of the pathology report generation datasets.}
The datasets span different disease settings, cohort sizes, and reporting conventions, enabling evaluation across heterogeneous report generation settings.}
\label{tab:dataset_comparison}
\begin{tabular}{lcccc}
\toprule
\textbf{Dataset} & \textbf{Train / Val / Test} & \textbf{Disease scope} &
\textbf{Report characteristics} & \textbf{Cohort size} \\
\midrule
TCGA-BRCA & 844 / 90 / 90 & Breast cancer &
Free-form clinical reports & 1,024 \\
HistAI & 9,980 / 1,248 / 1,248 & Multi-cancer &
Heterogeneous reports & 12,476 \\
REG-2025 & 5,924 / 740 / 741 & Multi-cancer &
Comparatively standardized reports & 7,405 \\
\bottomrule
\end{tabular}
\end{table*}

We evaluated SCOUT on three pathology report generation datasets that differ in disease composition, cohort size, and reporting characteristics:
TCGA-BRCA, HistAI, and REG-2025 (Table~\ref{tab:dataset_comparison}).
TCGA-BRCA provides a disease-specific breast cancer cohort, HistAI contains
a larger multi-cancer collection with heterogeneous report structures, and
REG-2025 provides a multi-cancer dataset with comparatively standardized
reports. Together, these datasets allow us to assess whether performance gains are consistently observed across distinct pathology report generation settings.

% We evaluated SCOUT on three pathology report generation datasets that differ in disease composition, dataset size, and reporting characteristics: TCGA-BRCA, HistAI, and REG-2025.  The three datasets  differ substantially in disease composition, institutional diversity, report homogeneity, and annotation noise, as summarized in Table~\ref{tab:dataset_comparison}. TCGA-BRCA represents a relatively narrow disease setting with breast cancer-specific diagnostic vocabulary, HistAI reflects a more heterogeneous multi-institutional setting with less standardized report structure, and REG-2025 provides a comparatively clean benchmark with consistent reporting conventions.
Evaluating across these complementary settings enables us to test whether the model remains robust when both visual distributions and linguistic styles vary. All comparison methods were trained using the same training, validation, and test splits, identical CONCHv1.5-derived patch features, and the same report preprocessing pipeline. We compared SCOUT with three recent WSI report generation methods: WSI-Caption~\cite{wsi_caption}, HistGen~\cite{histgen}, and Bi-Gen~\cite{bigen}.

Across the three datasets, SCOUT achieved the highest BLEU-1 through BLEU-4 and METEOR scores (Tables~\ref{tab:brca}--\ref{tab:histai}). On TCGA-BRCA, SCOUT obtained a BLEU-4 score of 0.1561 compared with 0.1394 for the strongest baseline, corresponding to a relative improvement of 12.0\%. METEOR increased from 0.1749 to 0.2038, a relative improvement of 16.5\%. SCOUT also achieved the highest ROUGE-L score on this dataset (0.3122).

Performance improvements were also observed on REG-2025, where SCOUT achieved BLEU-4, METEOR, and ROUGE-L scores of 0.7797, 0.5677, and 0.8739, respectively. Relative to the strongest competing method for each metric, these correspond to improvements of 6.0\%, 9.3\%, and 5.3\%. On HistAI, SCOUT improved BLEU-4 from 0.1281 to 0.1437 (12.2\%) and METEOR from 0.1472 to 0.1577 (7.1\%). Bi-Gen achieved a slightly higher ROUGE-L score on HistAI (0.3622 versus 0.3524), indicating that SCOUT did not uniformly outperform competing methods on every metric.

The improvements were therefore observed across datasets with substantially different report distributions rather than being restricted to a single benchmark. The larger gains in BLEU-3 and BLEU-4 indicate improved agreement with reference reports at the level of multi-token phrases and local linguistic structure.  However, these language-generation metrics do not directly measure whether clinically
important information is correct. We therefore separately assessed structured clinical fidelity using CRQS~\cite{pathreporteval}.

\begin{table}[t]
\centering
\caption{Performance on the TCGA-BRCA subset of PathText.}
\label{tab:brca}
\begin{tabular}{l|c|c|c|c|c|c}
\toprule
Method & BLEU-1 & BLEU-2 & BLEU-3& BLEU-4 & METEOR & ROUGE-L \\ %& Embedding Similarity  \\
\midrule
WSI-Caption   & 0.3621 & 0.2322 & 0.1513 & 0.1036 & 0.1521 & 0.2843  \\
HistGen & 0.4077 & 0.2594 & 0.1633 & 0.1060 & 0.1749 & 0.2670  \\
Bi-Gen  & 0.4155 & 0.2635 & 0.1794 & 0.1394 & 0.1581 & 0.2890  \\
\textbf{SCOUT} & \textbf{0.4363} & \textbf{0.3032} & \textbf{0.2024}& \textbf{0.1561} & \textbf{0.2038} & \textbf{0.3122} \\ %& \textbf{0.9551}\\
\bottomrule
\end{tabular}
\end{table}

\begin{table}[t]
\centering
\caption{Report generation performance on REG-2025 dataset. Best results are shown in bold.}
\label{tab:reg}
\begin{tabular}{l|c|c|c|c|c|c|c}
\toprule
Method & BLEU-1 & BLEU-2 & BLEU-3& BLEU-4 & METEOR & ROUGE-L & CRQS  \\
\midrule
WSI-Caption   & 0.8132 & 0.7775 & 0.7431 & 0.7153 & 0.5192 & 0.8146 & 0.6712  \\
HistGen & 0.8202 & 0.7840 & 0.7522 & 0.7251 & 0.4955  & 0.8116 & 0.6863\\
Bi-Gen  & 0.8248 & 0.7930 & 0.7605 & 0.7355 & 0.5041 & 0.8298 & 0.6985\\
\textbf{SCOUT} & \textbf{0.8651} & \textbf{0.8337} & \textbf{0.8050}& \textbf{0.7797} & \textbf{0.5677} & \textbf{0.8739} & \textbf{0.7376}\\
\bottomrule
\end{tabular}
\end{table}

\begin{table}[t]
\centering
\caption{Report generation performance on HistAI dataset. Best results are shown in bold.}
\label{tab:histai}
\begin{tabular}{l|c|c|c|c|c|c}
\toprule
Method & BLEU-1 & BLEU-2 & BLEU-3& BLEU-4 & METEOR & ROUGE-L \\ %& Embedding Similarity  \\
\midrule
WSI-Caption   & 0.2850 & 0.1930 & 0.1425 & 0.1114 & 0.1346 & 0.3478 \\ %& 0.901 \\
HistGen & 0.3111 & 0.2031 & 0.1678 & 0.1257 & 0.1466 & 0.3256 \\
Bi-Gen  & 0.3141 & 0.2177 & 0.1630 & 0.1281 & 0.1472 & \textbf{0.3622}  \\
\textbf{SCOUT} & \textbf{0.3598} & \textbf{0.2453} & \textbf{0.1830}& \textbf{0.1437} & \textbf{0.1577} & 0.3524 \\ %& \textbf{1.145}\\
\bottomrule
\end{tabular}
\end{table}

% We attribute this consistent behavior to two aspects of the proposed framework. First, the separation of patch-level, slide-level, and concept-level representations allows the model to preserve complementary information rather than collapsing all evidence into a single feature stream. Second, explicit concept conditioning provides a semantic anchor during generation, reducing sensitivity to superficial linguistic variation across datasets while improving focus on clinically relevant findings. Together, these results suggest that structured semantic priors and progressive contextual conditioning are effective mechanisms for improving cross-dataset generalization in pathology report generation.

\subsection{Clinically aligned evaluation with CRQS}
Standard language-generation metrics primarily quantify lexical similarity to reference reports and may not adequately capture clinically consequential errors such as omitted findings, unsupported statements, or discordant diagnostic attributes. We therefore additionally evaluated report quality using the Clinical Report Quality Score (CRQS), introduced in PathReportEval~\cite{pathreporteval}. PathReportEval provides a standardized evaluation framework for pathology report generation and introduces CRQS as a structured measure of clinical factual correctness. Rather than comparing reports directly at the token level, CRQS maps the
reference and generated reports to a predefined set of clinical attributes and evaluates four complementary aspects of report quality: Clinical Fact Coverage (CFC), which measures the proportion of reference clinical information recovered in the generated report; Key Information Recall (KIR), which measures recovery of diagnostically important attributes; Hallucination Rate (HR), which quantifies unsupported clinical information introduced in the generated report; and Clinical Discordance Score (CDS), which captures direct disagreement between generated and reference clinical attributes. These components are combined into an overall CRQS, with higher values indicating greater agreement with the clinically relevant content of the reference report. The complete definition, attribute extraction procedure, weighting scheme, and validation of CRQS are described in PathReportEval~\cite{pathreporteval}.

On the REG-2025 test set, SCOUT achieved a CRQS of 0.7376, compared with
0.6985 for Bi-Gen, 0.6863 for HistGen, and 0.6712 for WSI-Caption
(Table~\ref{tab:reg}). Relative to Bi-Gen, the strongest competing method,
SCOUT improved CRQS by 5.6\%. This result indicates that the improvements in BLEU, METEOR, and ROUGE-L were accompanied by improved agreement with
structured clinical information rather than reflecting lexical similarity
alone.

% \begin{table*}[t]
% \centering
% \caption{\textbf{Characteristics of datasets used for pathology report generation.}
% The datasets differ substantially in size, institutional diversity, annotation noise, and report structure, enabling a comprehensive evaluation of model robustness and generalization.}
% \label{tab:dataset_comparison}
% \begin{tabular}{l|c|c|c|c}
% \toprule
% \textbf{Dataset} & \textbf{Train / Val / Test} & \textbf{Diversity} & \textbf{Report Homogeneity} & \textbf{Annotation Noise} \\
% \midrule
% %TCGA  & 844 / 90 / 90 & $\sim$1041 & High & Medium & Medium--High & High \\
% TCGA-BRCA  & 844 / 90 / 90  & Low & Medium & High \\
% HistAI & 9980 / 1248 / 1248  & High & Low & Medium  \\
% REG 2025 & 5924 / 740 / 741  & High & High & Low  \\
% \bottomrule
% \end{tabular}
% \end{table*}

% For clinical relevance, we additionally report concept coverage, defined as the proportion of expert-annotated diagnostic concepts correctly mentioned in the generated report.

% \textbf{TCGA-BRCA dataset:}
% As shown in Table \ref{tab:brca}, our method demonstrates further gains relative to competing approaches, particularly in concept coverage, reflecting improved sensitivity to breast cancer–specific morphological patterns.

\subsection{Modality ablation analysis}

\begin{table}[t]
\centering
\caption{Ablation study performed on REG-2025 dataset.}
\label{tab:ablation}
\begin{tabular}{p{2cm}|p{3cm}|c|c|c|c|c|c}
\toprule
Features & Fusion method & BLEU-1 & BLEU-2 & BLEU-3& BLEU-4 & METEOR & ROUGE-L \\ %& Embedding Similarity  \\
\midrule
Patch  & -  & 0.8260 & 0.7886 & 0.7561 & 0.7281 & 0.5334 & 0.8341  \\
\hline
% Patch + slide & Plain concat & . & . & . & . & . & . \\
Patch + slide + concept  & Plain concat &  0.8454 & 0.8122 & 0.7820 & 0.7554 & 0.5551 & 0.8545  \\
\hline
Patch + slide + concept  & FiLM modulation + concat &  0.8511 & 0.8178 & 0.7878 & 0.7614 & 0.5511 & 0.8577  \\
\hline
Patch + slide + concept  & FiLM modulation + gated fusion &  0.8651 & 0.8337 & 0.8050 & 0.7797 & 0.5677 & 0.8739  \\
\bottomrule
\end{tabular}
\end{table}

We performed ablation experiments on REG-2025 dataset to examine the contributions of multimodal input representations, progressive FiLM-based context refinement, and adaptive gated fusion. All variants were evaluated using identical training and inference settings. The patch-only model achieved a BLEU-4 score of 0.7281, METEOR of 0.5334, and ROUGE-L of 0.8341 (Table~\ref{tab:ablation}). Providing patch-, slide-, and concept-level representations using static concatenation increased these scores to 0.7554, 0.5551, and 0.8545, respectively, demonstrating that additional global and semantic information contributes beyond local patch representations alone.

Replacing static feature integration with progressive FiLM conditioning provided a further improvement in BLEU-4 from 0.7554 to 0.7614 and ROUGE-L from 0.8545 to 0.8577. The effect on METEOR was smaller and was not monotonic (0.5551 versus 0.5511), indicating that iterative modulation does not improve every lexical metric independently. Nevertheless, the overall improvement in higher-order BLEU scores is consistent with a contribution from progressive interaction between local visual representations and contextual information.

Adding adaptive gated fusion on top of FiLM conditioning produced the strongest performance, increasing BLEU-4 from 0.7614 to 0.7797, METEOR from 0.5511 to 0.5677, and ROUGE-L from 0.8577 to 0.8739. Together, these ablations indicate that contextual representations and adaptive fusion provide complementary gains rather than the overall performance improvement arising from a single architectural component.

\textbf{Contribution of slide and concept features.}
Providing patch-, slide-, and concept-level representations using static
concatenation improved BLEU-4, METEOR, and ROUGE-L relative to the patch-only
model, indicating that the combined contextual representations provide
additional information beyond patch features alone.

\textbf{Effect of progressive FiLM conditioning.}
Replacing FiLM-based modulation with direct concatenation results in degradation across all evaluation metrics except METEOR, with the largest drops observed in higher-order BLEU scores. This suggests that iterative, depth-wise conditioning of patch representations is more effective than static feature fusion for capturing multi-scale dependencies. These findings support the hypothesis that progressive modulation enables better alignment between local visual evidence and global contextual signals.

\textbf{Effect of adaptive multimodal fusion.}
Substituting the gated fusion mechanism with static concatenation leads to performance declines across all metrics. This indicates that the relative importance of patch, slide, and concept features varies across tokens and cannot be captured by fixed fusion strategies. The proposed gating mechanism enables dynamic, token-level modulation of modality contributions, which appears critical for generating coherent and contextually grounded reports.

\subsection{Qualitative analysis of generated pathology reports}

\begin{figure}[t]
    \centering
    \includegraphics[width=1\linewidth]{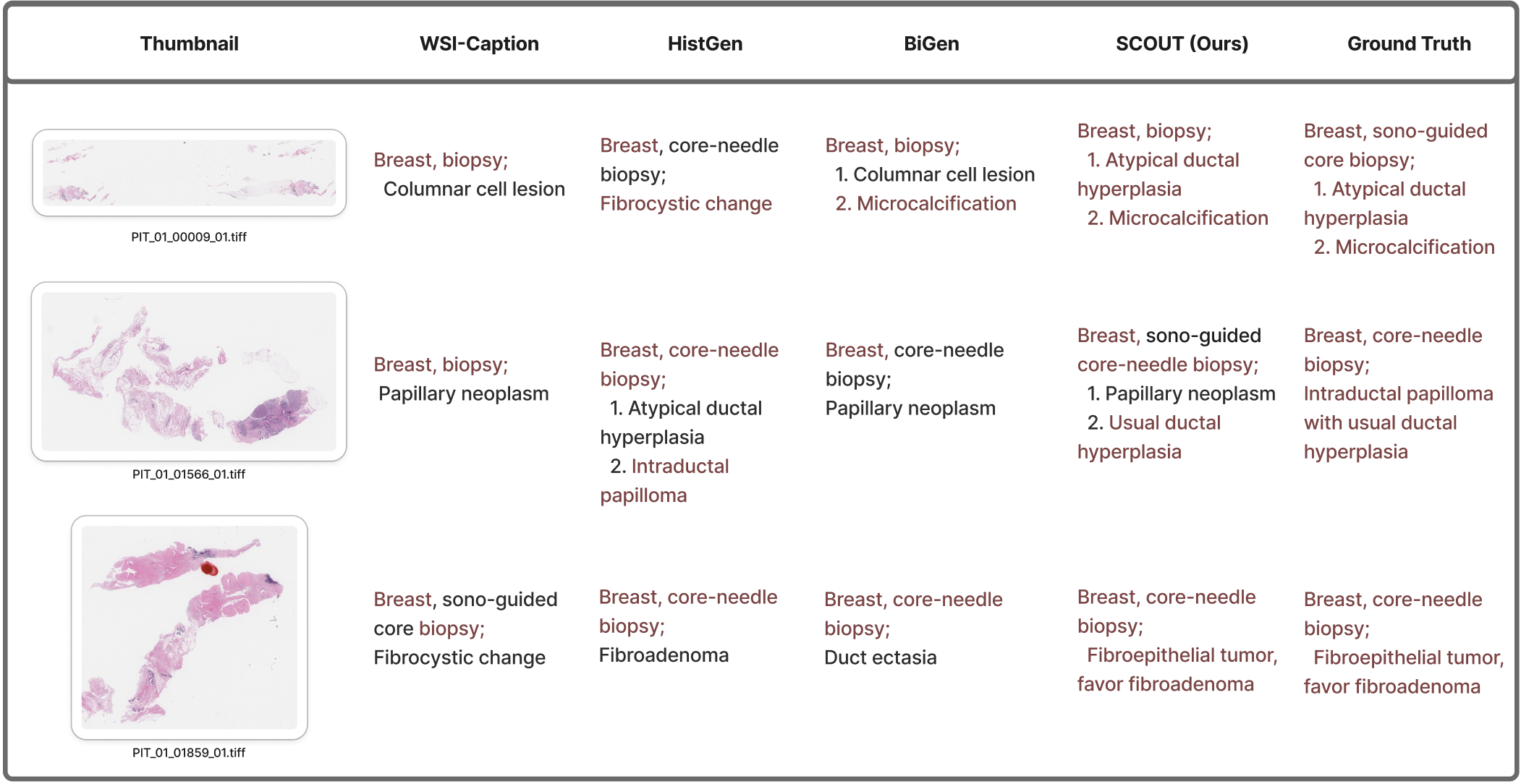}
    \caption{\textbf{Qualitative comparison of generated pathology reports.}
    Three test-set examples compare reports generated by WSI-Caption, HistGen, Bi-Gen, and SCOUT with the corresponding reference report. Highlighted text indicates overlapping or concordant diagnostic information. The examples illustrate differences in the coverage of diagnostically relevant findings and report-level consistency.}
    \label{fig:qualitative_results}
\end{figure}

Figure~\ref{fig:qualitative_results} compares reports generated by WSI-Caption, HistGen, Bi-Gen, and SCOUT with the corresponding reference reports for 3 example cases. Across the examples shown, all methods capture portions of the principal diagnosis, but differences are observed in the coverage and consistency of individual diagnostic findings. SCOUT reproduces a larger proportion of the highlighted reference content in these examples, while several baseline outputs omit or inconsistently describe individual findings.

These examples illustrate how the quantitative improvements can manifest at the report level but are not intended as an independent estimate of clinical performance. Clinical-content agreement across the full test cohort is instead quantified using CRQS (Table~\ref{tab:reg}).

\subsection{Multimodal fusion and modality contribution analysis}

\begin{figure}[t]
    \centering
    \includegraphics[width=0.6\linewidth]{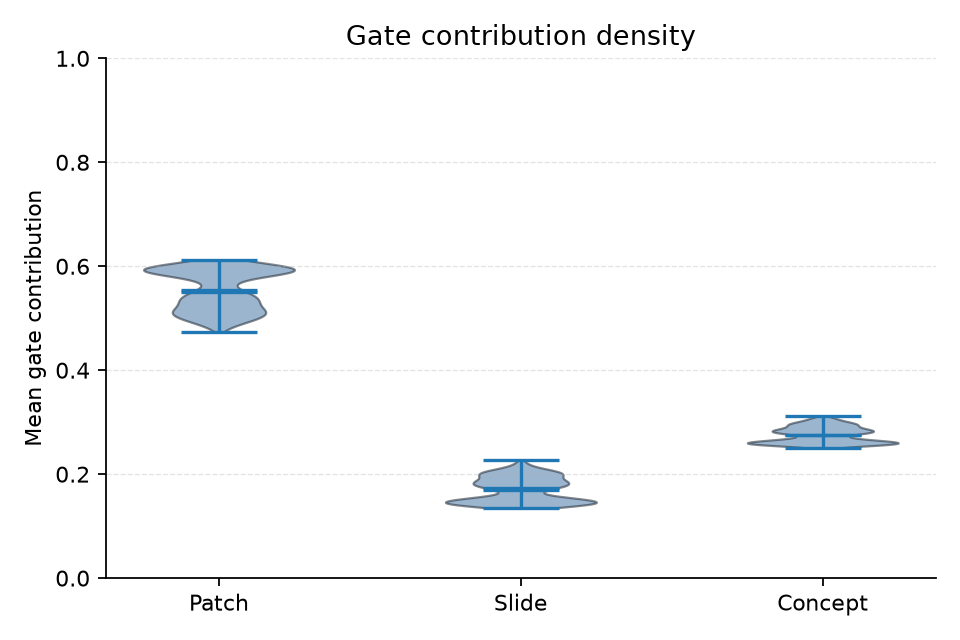}
    % \caption{\textbf{Distribution of learned modality contributions across the REG-2025 test cohort.}
    % Violin plots show case-level modality contributions obtained by averaging
    % final-layer feature-wise gate values across generated tokens, fusion heads,
    % and within-head feature dimensions. Patch representations receive the largest average contribution, followed by concept and slide representations. All three modalities retain non-zero contributions across the cohort, indicating that the learned fusion mechanism does not collapse onto a single representation stream. \pp{What does this plot mean?}}
    
    \caption{\textbf{Distribution of learned modality contributions across the REG-2025 test cohort.}
    For each test case, modality contribution was computed from the feature-wise gates of the final decoder layer by averaging gate values across all generated report tokens, fusion heads, and within-head feature dimensions. The resulting case-level values quantify the mean allocation assigned by the decoder to the patch, slide-conditioned, and concept-conditioned representation streams; gate weights are normalized across the three streams before aggregation. Violin width reflects the density of test cases at a given contribution value. Patch representations receive the largest average contribution across the cohort, concept-conditioned representations provide a substantial secondary contribution, and slide-conditioned representations receive a smaller but consistently non-zero contribution. The persistence of all three streams across cases indicates that the adaptive fusion mechanism does not collapse onto a single information source. These values describe model-internal modality allocation and should not be interpreted as causal measures of feature importance.
    % \pp{What does this plot mean?} \sk{updated caption}
    }
    \label{fig:gate_distribution}
\end{figure}

We next examined the learned fusion gates to characterize how patch-, slide-, and concept-level representations contribute during report generation. For each REG-2025 test case, feature-wise gate values from the final decoder
layer were averaged across generated report tokens, fusion heads, and
within-head feature dimensions to obtain one contribution per representation
stream. The resulting distributions were examined across the test cohort
(Fig.~\ref{fig:gate_distribution}).

Patch-level representations received the largest average gate contribution across the cohort, with values concentrated at approximately 0.5--0.6. Concept-level representations received the second-largest contribution, approximately 0.25--0.31, whereas slide-level representations contributed approximately 0.14--0.22. Thus, local morphological information remained the dominant representation during decoding, while explicit diagnostic concepts and global slide context provided complementary signals. Importantly, the learned gates did not collapse onto a single representation stream: patch, slide, and concept features retained non-zero contributions across the test cohort. The relatively consistent ordering of modality contributions suggests that SCOUT primarily bases generation on local visual evidence while incorporating semantic and whole-slide contextual information during decoding. These gate values should be interpreted as model-internal modality allocation rather than causal explanations of individual predictions.

\subsection{Case-level analysis of modality attention and gated fusion}

\begin{figure*}[t]
    \centering
    \includegraphics[width=\linewidth]{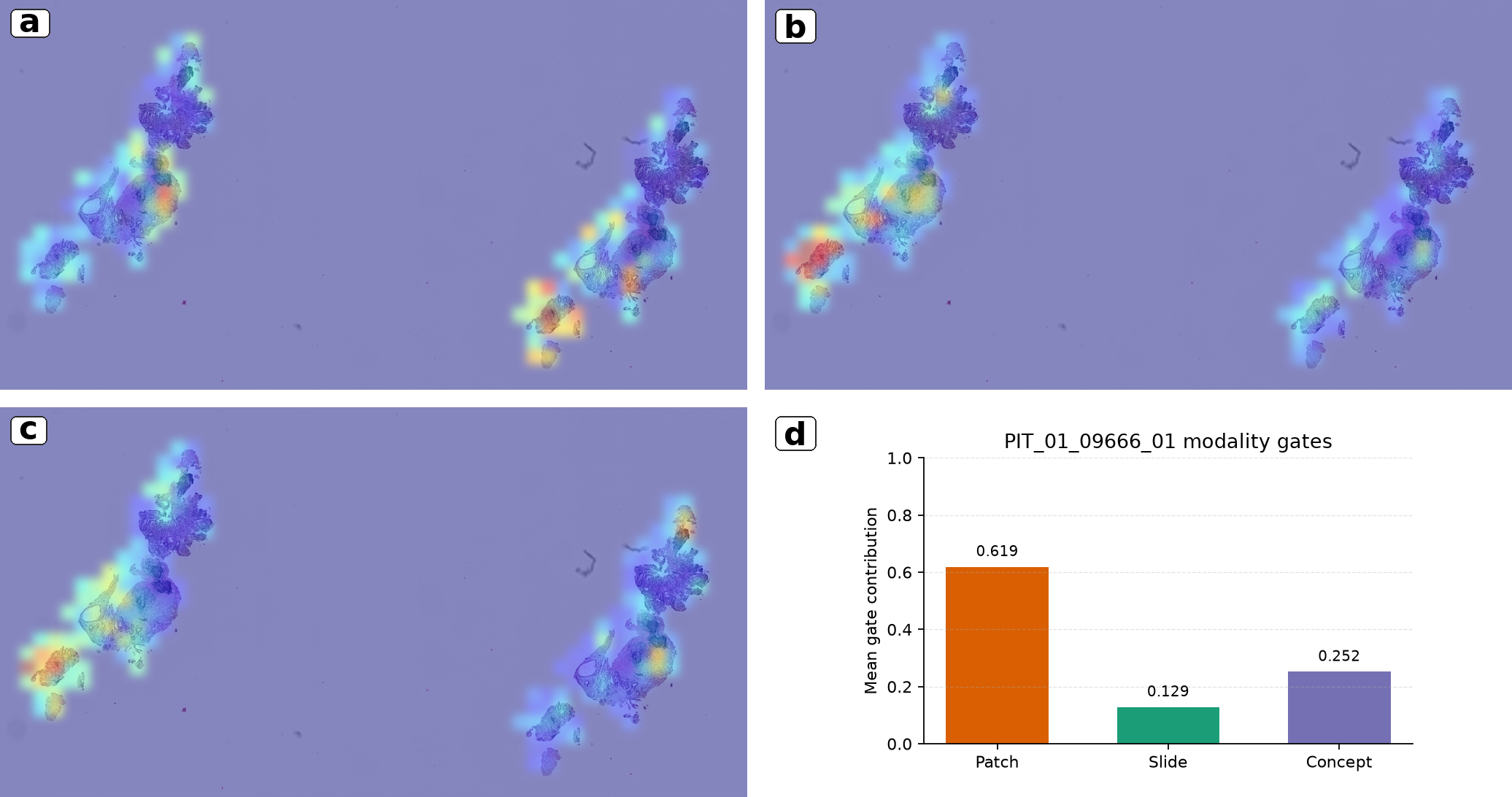}
    \caption{\textbf{Case-level visualization of modality-specific attention and gated fusion.}
    \textbf{a}, Patch-stream attention; \textbf{b}, slide-conditioned attention; and
    \textbf{c}, concept-conditioned attention for a representative REG-2025 test case.
    The three representation streams exhibit distinct attention
    distributions, illustrating differences in the information emphasized by each
    modality. 
    % \pp{How do we evaluate these? do we know where they are focusing on?}
    \textbf{d}, Mean decoder gate contributions for the same case, showing
    the relative contributions of patch-stream, slide-conditioned, and concept-conditioned representations during report generation. The visualization provides an illustrative view of
    modality use for an individual case (PIT\_01\_09666\_01) from REG-2025 dataset; cohort-level gate distributions are shown
    in Fig.~\ref{fig:gate_distribution}.}

%     \caption{\textbf{Case-level visualization of modality-specific attention andgated fusion.}
%     \textbf{a}, Patch-stream attention;
%     \textbf{b}, slide-conditioned attention; and
%     \textbf{c}, concept-conditioned attention for an illustrative REG 2025 test
%     case PIT_01
% }
    \label{fig:attention_gate_case}
\end{figure*}

To further illustrate how SCOUT uses complementary information sources during report generation, we examined modality-specific attention maps and decoder gate contributions for a representative case from the REG-2025 test set (Fig.~\ref{fig:attention_gate_case}). The figure combines patch-level, slide-level, and concept-level attention maps together with the corresponding gate contribution profile for the generated report.
Three additional test cases are provided in Appendix~\ref{app:additional_attention_cases}(Figs.~\ref{fig:attention_gate_case_03006}- \ref{fig:attention_gate_case_09073}).

% The patch-level attention map concentrates on localized regions with diagnostically relevant morphology, indicating that the model primarily relies on fine-grained tissue evidence when generating visually grounded findings. In contrast, the slide-level attention map exhibits a broader spatial distribution, consistent with the role of global context in summarizing tissue organization and specimen-level structure. The concept-level attention map highlights semantically relevant concept associations, suggesting that explicit histopathologic concepts provide an additional source of structured diagnostic information.

The patch-level attention map is comparatively localized, whereas the slide-conditioned attention map has a broader spatial distribution. The concept-conditioned stream shows a distinct attention pattern from both visual streams. The gate contribution profile for this case shows how these modality-specific signals are integrated during generation. Consistent with the cohort-level analysis (Fig.~\ref{fig:gate_distribution}), patch-level information contributes the largest share, while concept- and slide-conditioned representations provide complementary support. These report-level attention summaries illustrate that SCOUT draws on more than one representation stream during generation, though they do not establish that any particular region or modality caused a specific report statement.

These case-level examples should be interpreted as illustrative rather than exhaustive. Cohort-level modality contributions are summarized in Fig.~\ref{fig:gate_distribution}, whereas clinical report fidelity across the full test cohort is quantified using CRQS (Table~\ref{tab:reg}).

\section{Methods}
\subsection{Overview}

SCOUT is a multimodal Transformer architecture for pathology report generation that integrates evolving visual representations with recursively updated global and semantic context. Rather than treating visual understanding and language generation as sequential stages, the proposed approach enables continuous interaction between visual evidence, global context, and semantic concepts throughout the model.
The framework integrates three complementary sources of information:
(i) patch-level representations capturing fine-grained histological morphology,
(ii) slide-level embeddings encoding global tissue architecture, and
(iii) concept-level representations corresponding to clinically meaningful diagnostic attributes (Fig.~\ref{fig:architecture_overview}).
At a high level, these modalities are jointly leveraged within a unified architecture consisting of three stages: 
\begin{enumerate}
    \item \textbf{Multi-scale feature extraction and adaptation}: The pretrained patch, slide, and concept representations are transformed into a common latent space using modality-specific adaptation networks
    \item \textbf{Context-aware encoding with progressive concept conditioning}: A context-aware encoder
    maintains three evolving feature streams across network depth. At each encoder layer, the patch representation is updated using Transformer self-attention and spatial aggregation. The resulting representation is independently modulated by the current slide and concept states, and the corresponding context-conditioned representations are pooled to update the contextual states used at the following encoder layer. Representations from all encoder depths are subsequently combined using learned stream-specific weights.
    \item \textbf{adaptive multimodal decoding via contextual multi-head gating}: An autoregressive decoder independently attends to the patch, slide-conditioned, and concept-conditioned encoder outputs and combines them through adaptive multi-head gated fusion.
\end{enumerate}

This organization preserves local visual evidence, global contextual
information, and concept-conditioned information as separate representation streams until decoder fusion.

Unlike conventional conditional Transformers in which contextual information
remains fixed across network depth, SCOUT recursively updates its contextual
representations as the visual representation evolves. At each encoder layer,
the current patch representation is used to construct parallel
slide-conditioned and concept-conditioned feature streams. These streams are
subsequently pooled to update the slide and concept states used at the next
encoder depth. This produces a layer-wise co-evolution in which the visual
representation and its associated contextual states evolve jointly across the
encoder, while preserving a separate patch backbone.

% Unlike conventional conditional Transformers, where contextual information is provided as a fixed conditioning signal throughout the network, our encoder enables iterative co-evolution of visual and semantic representations. Specifically, patch features are progressively refined using slide-level context and diagnostic concepts, while the contextual representations themselves are updated after each encoder layer based on the refined visual evidence. This creates a feedback mechanism in which local morphology, global tissue context, and semantic concepts continuously influence one another during representation learning, allowing the model to progressively refine both visual features and their associated contextual interpretations. 

Together, these design choices provide a principled approach for integrating domain knowledge into vision–language models for computational pathology, emphasizing methodological transparency, adaptability, and clinical relevance.

\label{sec:Architecture overview}
\subsection{Architecture details}
% The proposed framework comprises four conceptual components [Figure \ref{fig:architecture_overview}]:
% \begin{itemize}
%     \item Modality-specific feature extractors that capture local histological detail, global slide context, and explicit diagnostic concepts.
%     \item Adaptive feature adaptation layers that project heterogeneous representations into a shared latent space
%     \item A context-aware encoder that progressively refines visual features under the influence of global context and semantic knowledge.
%     \item An adaptive multimodal decoder that integrates complementary information sources to generate coherent and clinically grounded reports
% \end{itemize}
% Each component is designed to reflect a distinct stage of diagnostic reasoning, from evidence extraction to contextual interpretation and narrative synthesis.

\subsubsection{Multi-scale visual feature extraction and adaptation}
\label{sec:Patch-Level Visual Features}
\paragraph{Patch-level visual features.}

Whole-slide images (WSIs) were divided into non-overlapping
$512\times512$ pixel tissue patches at $20\times$ magnification. Patch-level
representations were extracted using CONCHv1.5~\cite{conch}.

\noindent
\textbf{Notation:}
$d_p$, $d_t$, and $d_g$ denote the original embedding dimensions of frozen CONCHv1.5,
TITAN, and GECKO embeddings, respectively, while $d$ denotes
the shared latent dimensionality used throughout the encoder and decoder. $K$ denotes the number of diagnostic concepts. 
% \pp{why do these appear so late?} \sk{Moved up}

For a WSI containing $N$ retained tissue patches, the pretrained patch
representation is

\begin{equation}
\mathbf{P}_0 =
\{\mathbf{x}_1,\mathbf{x}_2,\ldots,\mathbf{x}_N\},
\qquad
\mathbf{x}_i\in\mathbb{R}^{d_p},
\end{equation}

where $d_p$ denotes the dimensionality of the CONCHv1.5 representation.

Patch features are transformed to the model dimensionality $d$ using a
three-layer nonlinear adapter,

\begin{equation}
\tilde{\mathbf{P}}
=
f_p(\mathbf{P}_0)
\in
\mathbb{R}^{B\times N\times d},
\end{equation}

where $B$ denotes batch size. 

Complete details of tissue segmentation, patch extraction, filtering,
feature extraction, and coordinate storage are provided in
Appendix~\ref{app:feature_extraction_details}.

% To capture fine-grained histological detail, we extract patch-level features from whole-slide images using CONCHv1.5\cite{conch}, a foundation model trained on large-scale histopathology data. WSIs are tiled into non-overlapping or weakly overlapping $512\times512$ patches at $20\times$ magnification, and each patch is embedded into a fixed-dimensional representation that encodes cellular morphology and local tissue structure.

These features are particularly effective at capturing morphological features like nuclear atypia, glandular structure, and cellular organization, but lack explicit global context. Addressing this limitation motivates the incorporation of complementary slide-level representations.

\paragraph{Learnable visual prompt.}
A learnable prompt token $\mathbf{p}\in\mathbb{R}^{d}$ is prepended to the adapted patch representation before encoding:

\begin{equation}
\mathbf{V}^{(0)}
=
[\mathbf{p};\tilde{\mathbf{P}}].
\end{equation}

The prompt token is randomly initialized and optimized jointly with the
report-generation model. The resulting visual sequence is padded as required
for spatial processing by PAM; we denote its resulting length by $M$,

\begin{equation}
\mathbf{V}^{(0)}
\in
\mathbb{R}^{B\times M\times d}.
\end{equation}

\label{sec:Slide-Level Global Context}
% \paragraph{Slide-level global context.}

\paragraph{Slide-level contextual representations}

Global contextual information is represented using complementary pretrained
slide embeddings derived from TITAN and the deep representation of
GECKO~\cite{gecko}.

Let $\mathbf{s}_{T}\in\mathbb{R}^{d_t},\mathbf{s}_{G}\in\mathbb{R}^{d_g}$ denote the TITAN and GECKO deep representations, respectively. Each is first
processed by an independent adapter,

\begin{align}
\tilde{\mathbf{s}}_{T}
&=
f_T(\mathbf{s}_{T})
\in\mathbb{R}^{d},\\
\tilde{\mathbf{s}}_{G}
&=
f_G(\mathbf{s}_{G})
\in\mathbb{R}^{d}.
\end{align}

The adapted representations are concatenated and passed through a dedicated
slide-fusion network,

\begin{equation}
\mathbf{s}^{(0)}
=
f_s
\left(
[\tilde{\mathbf{s}}_T;\tilde{\mathbf{s}}_G]
\right)
\in\mathbb{R}^{d}.
\end{equation}

The TITAN and GECKO adapters each use a learnable multi layer perceptron  with ReLU
activations, dropout, and output LayerNorm.  The resulting
$\mathbf{s}^{(0)}$ provides the initial slide-context state of the
context-aware encoder. Further details of TITAN and GECKO feature extraction are provided in
Appendix~\ref{app:slide_features_construction}.

% To encode global tissue architecture and long-range spatial organization, we extract slide-level embeddings using TITAN, a whole-slide representation model designed to summarize large-scale histological patterns. Unlike patch-level features, these embeddings capture contextual information such as tissue composition, spatial heterogeneity, and global morphological trends. These embeddings are further enriched using GECKO deep slide level representations which provide complimentary global information; further implementation details are provided in ~\ref{app:slide_features_construction}. Each slide is represented as a single vector: $\mathbf{s} \in \mathbb{R}^{d_s}$

% \begin{equation}
% \mathbf{s} \in \mathbb{R}^{D_s}.
% \end{equation}

These features summarize slide-wide characteristics such as architectural patterns and staining distributions, etc. While such representations are insufficient for fine-grained description on their own, they provide crucial contextual signals that guide the interpretation of local visual evidence.

\label{sec:Concept-Level Semantic Features}
\paragraph{Concept-level semantic features.}
In addition to purely visual representations, we incorporate explicit diagnostic concepts derived through a semi-automated pipeline combining large language model (LLM)–assisted extraction and expert curation using GECKO~\cite{gecko}. Candidate concept descriptions were derived from pathology reports and domain-specific textual resources using GPT-5.3. These candidates were subsequently reviewed by domain experts to remove redundant, ambiguous, or
clinically inappropriate concepts and to standardize terminology. Reports from
the held-out test cohorts were not used during concept-vocabulary construction. The resulting concept set therefore reflects both data-driven discovery and domain expertise, providing a structured semantic representation aligned with routine diagnostic practice. For each WSI, GECKO produces a concept-level feature vector indexed by the
fixed vocabulary,

\begin{equation}
\mathbf{q}
=
[q_1,q_2,\ldots,q_K]
\in
\mathbb{R}^{K}.
\end{equation}

The concept feature vector is transformed to the common model dimensionality
using a modality-specific adapter,

\begin{equation}
\mathbf{c}^{(0)}
=
f_c(\mathbf{q})
\in
\mathbb{R}^{d}.
\end{equation}

The concept vocabulary is fixed after construction and the same concept definitions are used during training and inference. At inference time, no text from the target pathology report is used to construct the concept representation; the case-specific concept features are derived from the WSI using GECKO. Additional details of concept generation, expert curation, and GECKO concept-feature construction are provided in Appendix~\ref{app:feature_extraction_details}.

% Each case is associated with a concept embedding: $\mathbf{C} = \{\mathbf{c}_1, \mathbf{c}_2, \dots, \mathbf{c}_K\}, 
% \quad \mathbf{c}_k \in \mathbb{R}^{d_c},$ where $K$ denotes the number of concepts. These embeddings are extracted using CONCHv1 encoder\cite{conch} and encode clinically meaningful attributes such as architectural patterns and cytological abnormalities. The concept feature extraction process is further described in detail in appendix~\ref{app:feature_extraction_details}.

These features represent the diagnostic workflow of human pathologists, where known disease entities and morphological patterns shape visual interpretation rather than being inferred only after observation.

\subsubsection{Context-Aware Encoder with Progressive Modulation}
\label{sec:Context aware Encoder}
WSIs exhibit extreme heterogeneity: diagnostically relevant regions are sparse, context-dependent, and often ambiguous when viewed in isolation. While patch-level representations capture local morphology, their interpretation depends critically on global slide context (e.g., tumor burden, architectural patterns) and explicit diagnostic concepts (e.g., Gleason patterns, necrosis). We therefore design a context-aware encoder in which the evolving patch representation is used at every depth to construct parallel slide-conditioned and concept-conditioned feature streams. These contextual streams are recursively updated across encoder layers, enabling the contextual representation associated with a WSI to adapt as increasingly transformed visual features become available.

\label{sec:Encoder structure}
\paragraph{Layer-wise patch encoding}

The context-aware encoder contains $L$ Transformer layers. At encoder depth
$\ell$, the visual representation from the preceding layer is first
normalized and passed through a Transformer encoder block containing
multi-head self-attention and a position-wise feed-forward network with
residual connections:

\begin{equation}
\mathbf{U}^{(\ell)}
=
\operatorname{EncoderLayer}_{\ell}
\left(
\operatorname{Norm}
\left(
\mathbf{P}^{(\ell-1)}
\right)
\right),
\end{equation}

with $\mathbf{P}^{(0)}=\mathbf{V}^{(0)}$. The encoder implementation uses pre-normalized residual sublayers for both self-attention and feed-forward transformations.

A spatial aggregation module $\mathrm{PAM}(\cdot)$  is subsequently applied, that enhances local contextual interactions\cite{wsi_caption}:

\begin{equation}
\mathbf{P}^{(\ell)}
=
\operatorname{PAM}_{\ell}
\left(
\mathbf{U}^{(\ell)}
\right).
\end{equation}
This operation captures spatial and morphological relationships among image patches; additional details of PAM module are provided in Appendix~\ref{app:pam}. The resulting patch representation is passed directly to the next Transformer layer and is also used to construct the two context-conditioned streams.

\label{sec:FiLM-Based Contextual Modulation}
\paragraph{FiLM-based contextual modulation.}
To incorporate global slide information and concept-level semantics, we apply feature-wise linear modulation (FiLM\cite{film}) to patch representations. Given a conditioning vector $\mathbf{z} \in \mathbb{R}^{B \times d}$ (either slide or concept embedding), FiLM produces feature-wise scaling and shifting parameters. For layer $\ell$, we compute modulation parameters $\beta$ and $\gamma$ separately for both slide and concept features:

% \begin{equation}
% [\boldsymbol{\gamma}^{(\ell)}, \boldsymbol{\beta}^{(\ell)}] = g(\mathbf{z}^{(\ell)}),
% \end{equation}

\begin{equation}
[
\boldsymbol{\gamma}_{m}^{(\ell)},
\boldsymbol{\beta}_{m}^{(\ell)}
]
=
g_{m}^{(\ell)}
\left(
\mathbf{z}_{m}^{(\ell-1)}
\right),
\end{equation}

where $m\in\{s,c\}$, $g_m^{(\ell)}(\cdot)$ is a learnable multilayer perceptron, and $\boldsymbol{\gamma}_m^{(\ell)},\boldsymbol{\beta}_m^{(\ell)}\in\mathbb{R}^{B\times d}$. These parameters are applied to the patch features as:
% \begin{equation}
% \mathrm{FiLM}(\mathbf{P}, \mathbf{z}) = 
% \mathrm{LayerNorm}\left(
% \mathbf{P} \odot \left(1 + \alpha \boldsymbol{\gamma} \right) + \boldsymbol{\beta}
% \right),
% \end{equation}

\begin{equation}
\operatorname{FiLM}_{m}^{(\ell)}
(\mathbf{P},\mathbf{z}_{m})
=
\operatorname{LayerNorm}
\left[
\mathbf{P}
\odot
\left(
1+\alpha\boldsymbol{\gamma}_{m}^{(\ell)}
\right)
+
\boldsymbol{\beta}_{m}^{(\ell)}
\right],
\end{equation}

where $\odot$ denotes element-wise multiplication and $\alpha$ controls the strength of modulation. Slide and concept modulation are applied in parallel:

\begin{align}
\mathbf{S}^{(\ell)}
&=
\operatorname{FiLM}_{s}^{(\ell)}
\left(
\mathbf{P}^{(\ell)},
\mathbf{s}^{(\ell-1)}
\right),
\mathbf{C}^{(\ell)}
=
\operatorname{FiLM}_{c}^{(\ell)}
\left(
\mathbf{P}^{(\ell)},
\mathbf{c}^{(\ell-1)}
\right).
\end{align}

Both outputs retain the spatial token structure of the patch representation, $\mathbf{S}^{(\ell)}, \mathbf{C}^{(\ell)} \in \mathbb{R}^{B\times M\times d}$. The two outputs therefore provide spatially resolved representations of the same visual tokens conditioned independently on global slide context and diagnostic concept information.

% This enables the encoder to emphasize features aligned with known pathological concepts while suppressing visually plausible but clinically irrelevant patterns.

\subsubsection{Iterative Co-evolution of Patch, Slide and Concept Representations}

The slide and concept contextual states are updated after every encoder layer
rather than remaining fixed throughout the network. The context-conditioned
visual representations are mean-pooled over the visual-token dimension:

\begin{align}
\mathbf{s}^{(\ell)}
&=
\operatorname{MeanPool}
\left(
\mathbf{S}^{(\ell)}
\right),
\mathbf{c}^{(\ell)}
=
\operatorname{MeanPool}
\left(
\mathbf{C}^{(\ell)}
\right).
\end{align}

After each encoder layer, the slide-conditioned and concept-conditioned patch representations are pooled to update their respective contextual states, which subsequently serve as conditioning signals for the following encoder layer. The updated features  are $\{
\mathbf{P}^{(\ell)},
\mathbf{s}^{(\ell)},
\mathbf{c}^{(\ell)}
\}$.

Consequently, contextual information is not merely injected into the visual stream but is itself progressively refined based on the evolving visual evidence. This yields recursively updated slide and concept contextual states for the next layer. The patch stream evolves through successive Transformer and PAM operations, whereas the slide and concept states are recursively recomputed from their corresponding conditioned visual representations. The updated contextual states subsequently condition their respective FiLM branches at the following encoder layer. We refer to this layer-wise evolution of the visual and contextual states as iterative co-evolution.

% where $\mathrm{Pool}(\cdot)$ denotes mean aggregation across patch tokens. This creates a feedback loop in which global and semantic representations are iteratively refined based on evolving visual evidence.

\paragraph{Depth-weighted feature aggregation}

Rather than using only the final encoder layer, SCOUT aggregates representations across encoder depth independently for each stream. For $m\in\{\mathrm{patch},\mathrm{slide},\mathrm{concept}\}$, the normalized weight assigned to layer $\ell$ is $w_{\ell}^{(m)}
=
\frac{
\exp(\theta_{\ell}^{(m)})
}{
\sum_{j=1}^{L}
\exp(\theta_{j}^{(m)})
}$, where $\theta_{\ell}^{(m)}$ is a learnable scalar parameter. The layer-specific representations are

\begin{align}
\mathbf{H}^{(\ell,\mathrm{patch})}
=
\mathbf{P}^{(\ell)},
\mathbf{H}^{(\ell,\mathrm{slide})}
=
\mathbf{S}^{(\ell)},
\mathbf{H}^{(\ell,\mathrm{concept})}
=
\mathbf{C}^{(\ell)}.
\end{align}

For each stream, the final encoder representation is $\mathbf{F}^{(m)} =
\operatorname{Norm}_{m}
\left[
\sum_{\ell=1}^{L}
w_{\ell}^{(m)}
\mathbf{H}^{(\ell,m)}
\right]$. This produces three spatially resolved encoder outputs,

\begin{equation}
\mathbf{F}^{(\mathrm{patch})},
\mathbf{F}^{(\mathrm{slide})},
\mathbf{F}^{(\mathrm{concept})}
\in
\mathbb{R}^{B\times M\times d},
\end{equation}

corresponding to the patch, slide-conditioned, and concept-conditioned streams. Additional implementation details for the encoder are provided in Appendix~\ref{app:encoder_details}.

\subsubsection{Multimodal Decoder with Adaptive 3-Way Gated Fusion}
To generate clinically coherent reports, the decoder must integrate heterogeneous sources of information, including fine-grained patch features, global slide context, and diagnostic concept representations. Rather than relying on early fusion or simple concatenation, we design a decoder that explicitly models interactions between these modalities through parallel cross-attention pathways and adaptive gated fusion.

\paragraph{Autoregressive decoding.}

Report tokens are embedded into the model dimensionality $d$ and augmented
with sinusoidal positional encodings. Given the preceding target tokens,
causal self-attention constructs a decoder representation while preventing
access to future tokens.

For decoder layer $j$, we denote the output of the causal self-attention
sublayer as

\begin{equation}
\mathbf{D}^{(j)}
=
\operatorname{SelfAttnBlock}
\left(
\mathbf{D}^{(j-1)}
\right).
\end{equation}

All decoder attention and feed-forward modules are implemented using
pre-normalized residual sublayers.

% Given a sequence of previously generated tokens $\mathbf{y}_{<t}$, the decoder produces hidden representations using causal self-attention: $\mathbf{H}_t = \mathrm{SelfAttn}(\mathbf{y}_{<t})$

% ons depend only on past tokens.

\paragraph{Modality-specific cross-attention.}

Each decoder layer contains three independent cross-attention modules. The
same decoder state attends separately to the patch, slide-conditioned, and
concept-conditioned encoder representations:

\begin{equation}
\tilde{\mathbf{D}}_{m}^{(j)}
=
\operatorname{CrossAttnBlock}_{m}
\left(
\mathbf{D}^{(j)},
\mathbf{F}^{(\mathrm{m})}
\right), \qquad m \in \{\mathrm{patch}, \mathrm{slide}, \mathrm{concept}\}
\end{equation}

Each cross-attention block includes normalization, multi-head scaled
dot-product attention, dropout, and a residual connection. The three branches
use separate cross-attention parameters. This design allows each modality to contribute complementary information: patch features capture local morphology, slide features encode global tissue context, and concept features represent high-level diagnostic semantics. The corresponding attention tensors are retained independently for subsequent modality-attention analysis.

\paragraph{Limitations of naive fusion.}
A straightforward combination of these representations (e.g., summation or concatenation) assumes equal importance across modalities and tokens, which is unrealistic in pathology. The relevance of each modality varies dynamically depending on the linguistic context (e.g., cellular descriptions vs. diagnostic summaries), necessitating an adaptive fusion mechanism.

\paragraph{Feature-wise multi-head gated fusion}

The three cross-attended representations are combined using an adaptive
multi-head gated fusion module. The modality streams are first independently
projected:

\begin{align}
\mathbf{R}_{p}
=
W_p\tilde{\mathbf{D}}_{p},
\mathbf{R}_{s}
=
W_s\tilde{\mathbf{D}}_{s},
\mathbf{R}_{c}
=
W_c\tilde{\mathbf{D}}_{c}.
\end{align}

The current decoder representation and the three cross-attended
representations are concatenated,
$
\mathbf{Q}
=
[
\mathbf{D};
\tilde{\mathbf{D}}_{p};
\tilde{\mathbf{D}}_{s};
\tilde{\mathbf{D}}_{c}
]$, and processed by the gating network $\mathbf{Z}=\phi(\mathbf{Q}),$
where $\phi(\cdot)$ is an MLP with ReLU activations and dropout. Let $H$ denote the number of fusion heads and $d_h=d/H$ the dimensionality of each head. The gating output is reshaped as
$
\mathbf{Z}
\in
\mathbb{R}^{B\times T\times H\times3\times d_h},
$ where $T$ denotes the decoder sequence length and the fourth dimension indexes
the patch, slide-conditioned, and concept-conditioned streams. Unlike a scalar gate for each modality and head, SCOUT predicts a feature-wise gate within every head. For output token $t$, head $h$, modality $m$, and head feature dimension $r$, the normalized gate weight is

\begin{equation}
\alpha_{t,h,m,r}
=
\frac{
\exp
\left(
z_{t,h,m,r}/\tau
\right)
}{
\sum_{m'}
\exp
\left(
z_{t,h,m',r}/\tau
\right)
},
\end{equation}

where $\tau$ is a learnable temperature parameter initialized to 1. The
softmax is applied over the three modalities and therefore satisfies $\sum_m \alpha_{t,h,m,r}=1$ for every token, head, and feature dimension. After reshaping the projected modality representations into $H$ heads, multimodal fusion is performed element-wise within each head:

\begin{equation}
\mathbf{R}_{t,h,r}
=
\alpha_{t,h,p,r}\mathbf{R}_{p,t,h,r}
+
\alpha_{t,h,s,r}\mathbf{R}_{s,t,h,r}
+
\alpha_{t,h,c,r}\mathbf{R}_{c,t,h,r}.
\end{equation}

The fused heads are concatenated and projected to $d$ dimensions. The fusion
module incorporates a residual connection and LayerNorm followed by an
additional feed-forward transformation. The complete fusion operation is
itself applied within a residual decoder sublayer, after which a standard
position-wise feed-forward sublayer completes the decoder layer. The decoder therefore follows the sequence

\begin{equation}
\mathrm{causal\ self\!-\!attention}
\rightarrow
\mathrm{three\ parallel\ cross\!-\!attentions}
\rightarrow
\mathrm{adaptive\ gated\ fusion}
\rightarrow
\mathrm{feed\!-\!forward}.
\end{equation}

Attention maps and fusion weights are retained at every decoder layer. The
final decoder output is normalized and projected to the report vocabulary to
obtain next-token probabilities. Additional implementation details for the decoder are provided in Appendix~\ref{app:decoder_details}.

% \paragraph{Interpretability and practical advantages.}
% The learned gating weights $\boldsymbol{\alpha}$ provide an explicit measure of modality importance at each token and attention head, enabling fine-grained analysis of how visual evidence and diagnostic concepts contribute to generated text.

% This design allows the model to dynamically shift between modalities, supporting both descriptive (visual grounding) and explanatory (concept-driven reasoning) components of pathology reports. Decoding proceeds autoregressively until an end-of-sequence token is generated.

\paragraph{Modality-attribution signals.}
The feature-wise gate weights provide model-internal estimates of the relative
allocation assigned to the patch, slide-conditioned, and concept-conditioned
streams during generation. We retain these weights, together with
modality-specific cross-attention tensors, for the analyses described below.
These quantities are used to characterize model behavior and are not
interpreted as causal explanations of individual predictions.

% This decoder design provides several advantages. First, the adaptive gating mechanism enables dynamic modulation of modality contributions, allowing the model to shift between fine-grained visual grounding and high-level semantic reasoning depending on the linguistic context. Second, by separating modality-specific cross-attention pathways and combining them through learned fusion, the model avoids over-reliance on any single information source, improving robustness. Third, the learned gating weights offer interpretable signals that reflect the relative influence of visual evidence and diagnostic concepts at each decoding step, enabling analysis of reasoning patterns. Finally, the integration of concept representations as primary conditioning signals—rather than auxiliary supervision—encourages alignment between generated text and clinically meaningful semantic structures.

\subsection{Training Objectives}

SCOUT is optimized using autoregressive report-generation loss together with
gate entropy regularization.

For reference report
$\mathbf{y}=(y_1,\ldots,y_T)$
and multimodal input $\mathbf{X}$, the negative log-likelihood objective is

\begin{equation}
\mathcal{L}_{\mathrm{NLL}}
=
-
\sum_{t=1}^{T}
\log
p
\left(
y_t
\mid
y_{<t},
\mathbf{X}
\right).
\end{equation}

\subsubsection{Gate entropy regularization}

The gating mechanism defines a normalized distribution over the three
modalities for every token, fusion head, and within-head feature dimension.
To encourage selective modality allocation, we minimize the entropy of these
distributions:

\begin{equation}
\mathcal{L}_{\mathrm{gate}}
=
\mathbb{E}_{t,h,r}
\left[
-
\sum_m
\alpha_{t,h,m,r}
\log
\left(
\alpha_{t,h,m,r}+\epsilon
\right)
\right],
\end{equation}

where $\epsilon$ is included for numerical stability.

This regularization encourages concentrated modality allocations. In combination with the report-generation objective, it allows different modalities to dominate at different stages of report generation while retaining complementary information when useful.

The complete objective is

\begin{equation}
\mathcal{L}
=
\mathcal{L}_{\mathrm{NLL}}
+
\lambda_g
\mathcal{L}_{\mathrm{gate}},
\end{equation}

where $\lambda_g$ controls the strength of gate regularization.

\subsection{Modality contribution and attention analysis}

\subsubsection{Cohort-level modality contributions}

The feature-wise gate tensors returned by the decoder were retained during
report generation. To obtain a single contribution for modality $m$ in test
case $i$, fusion weights from the final decoder layer were averaged across
generated report tokens, fusion heads, and within-head feature dimensions:

\begin{equation}
\bar{\alpha}_{i,m}
=
\frac{1}{T_i H d_h}
\sum_{t=1}^{T_i}
\sum_{h=1}^{H}
\sum_{r=1}^{d_h}
\alpha_{i,t,h,m,r}.
\end{equation}

The distributions of
$\bar{\alpha}_{i,m}$
across the REG 2025 test cohort were used to construct the violin plots in
Fig.~\ref{fig:gate_distribution}.

The case-level modality-contribution bar plot in
Fig.~\ref{fig:attention_gate_case} was calculated using the same aggregation
for the illustrated test case.

\subsubsection{Modality-specific spatial attention analysis}

The decoder returns cross-attention tensors independently for the patch,
slide-conditioned, and concept-conditioned streams at every decoder layer.
For the case-level visualization, attention weights from the final decoder
layer were used.

For modality $m$, attention was averaged across decoder attention heads and
generated report tokens:

\begin{equation}
A_m(n)
=
\frac{1}{T H_a}
\sum_{t=1}^{T}
\sum_{h=1}^{H_a}
A^{(\mathrm{last})}_{t,h,m,n},
\end{equation}

where $H_a$ is the number of cross-attention heads and $n$ indexes visual
tokens.

Attention corresponding to non-tissue/padding tokens was excluded before
visualization, and the remaining patch-level values were mapped to their
stored WSI coordinates. This produced separate spatial attention maps for the
patch, slide-conditioned, and concept-conditioned streams shown in
Fig.~\ref{fig:attention_gate_case}. Because attention is averaged across the
entire generated sequence, these maps summarize report-level spatial
attention rather than attribution to an individual generated word.

% The model is trained to generate pathology reports while maintaining alignment with diagnostic concepts and structured multimodal reasoning. To this end, we optimize a composite objective consisting of language modeling, concept supervision, and attention regularization. Our model is trained using the standard negative log-likelihood objective:

% \begin{equation}
% \mathcal{L}_{\text{NLL}}
% =
% - \sum_{t=1}^{T}
% \log p(y_t \mid y_{<t}, \mathbf{X}),
% \end{equation}

% where $\mathbf{X}$ denotes the multimodal input comprising patch, slide, and concept features.

% The decoder combines modality-specific representations using a learnable gating mechanism that produces weights over patch, slide, and concept features. To encourage interpretable and selective modality usage, we introduce a soft sparsity regularization on the gating weights based on entropy minimization:

% \begin{equation}
% \mathcal{L}_{\text{gate}}
% =
% \mathbb{E}_{t,h}
% \left[
% - \sum_{m}
% w_{t,h,m} \log (w_{t,h,m} + \epsilon)
% \right],
% \end{equation}

% where $w_{t,h,m}$ denotes the gating weight for modality $m$ at decoding step $t$ and attention head $h$, and $\epsilon$ is a small constant for numerical stability.

% This regularization encourages the model to make confident, but not strictly exclusive, modality selections, allowing different modalities to dominate at different stages of report generation while still supporting complementary information when needed.

% \paragraph{Overall objective.}

\section{Experimental Setup and Evaluation Protocol}

\subsection{Datasets and Splits}
% All experiments were conducted on curated whole-slide image datasets paired with pathology reports. Slides were partitioned into training, validation, and test sets at the case level. Unless otherwise stated, validation data were used exclusively for model selection and hyperparameter tuning, while the test set was held out until final evaluation. We evaluate the proposed framework on three datasets TCGA-BRCA (from PathText), HistAI, and REG-2025 selected to cover complementary levels of disease specificity, institutional diversity, report homogeneity, and annotation noise.

We evaluated SCOUT on three datasets that differ in disease composition,
cohort size, and report structure: TCGA-BRCA, HistAI, and REG-2025
(Table~\ref{tab:dataset_comparison}). TCGA-BRCA provides a disease-specific
breast cancer cohort, HistAI contains a larger multi-cancer collection with
heterogeneous report formats, and REG-2025 contains comparatively standardized
reports. Together, these datasets provide complementary pathology report
generation settings for evaluating consistency of model performance.

All experiments were conducted on whole-slide image datasets paired with pathology reports. For each dataset, we constructed fixed training, validation, and test partitions using an approximately 80/10/10 case-level split. These splits were created once before model development and were held fixed across all experiments and comparison methods. Validation data were used for model selection and hyperparameter tuning, whereas the test partitions were reserved for final evaluation.

The same official split files used in the experiments are released with the project repository to facilitate exact reproduction of the reported results. The resulting train/validation/test sizes are summarized in Table~\ref{tab:dataset_comparison}.

% Table~\ref{tab:dataset_comparison} summarizes the key differences among these datasets and provides context for the quantitative results in Section~4.1. In particular, TCGA-BRCA represents a disease-specific setting with relatively lower diversity but noisier reports, HistAI reflects a more heterogeneous multi-institutional setting with less standardized reporting, and MICCAI REG provides a comparatively cleaner and more homogeneous benchmark. Evaluating across these three settings allows us to assess not only absolute report-generation quality, but also the robustness of the proposed method under varying reporting conventions and data characteristics.

\paragraph{PathText (TCGA-BRCA subset).} We use the breast cancer subset of PathText, a publicly available pathology report generation dataset derived from The Cancer Genome Atlas (TCGA), consisting of whole-slide histopathology images paired with corresponding diagnostic reports~\cite{wsi_caption}. Relative to the other datasets, this subset is more disease-specific and contains free form text with greater variation in report content and structure. This dataset also sometimes includes information beyond what is directly observable from the WSI, including clinical history and gross measurements. Consequently, some reference-report content is not reliably recoverable from image features alone. We therefore include this dataset as a more challenging setting to evaluate if models can produce accurate reports despite these issues.

\paragraph{HistAI.} The HistAI dataset provides high-resolution histopathology slides with associated structured and free-text annotations curated for clinical AI research~\cite{histai_dataset}. As summarized in Table~\ref{tab:dataset_comparison}, HistAI is more institutionally diverse and less homogeneous in reporting structure than REG-2025, making it a useful test bed for evaluating robustness to inter-site variability and differences in reporting style. We therefore use HistAI to assess performance under greater visual and linguistic heterogeneity.

\paragraph{REG-2025 Challenge Dataset.} The Report Generation challenge dataset was part of MICCAI 2025 and includes histopathological images paired with expert-authored diagnostic reports, designed specifically to benchmark automated report generation methods~\cite{reg2025_challenge}. Compared with TCGA-BRCA and HistAI, this dataset is comparatively cleaner and more homogeneous in report style, as indicated in Table~\ref{tab:dataset_comparison}. We include it to measure performance in a standardized evaluation setting and to determine whether the proposed method remains effective even when reporting conventions are more uniform.

\subsection{Image Preprocessing and Patch Sampling}
Whole-slide images (WSIs) were tessellated into non-overlapping
$512\times512$ pixel patches at $20\times$ magnification. Patch-level visual
features were extracted using the pretrained CONCHv1.5 encoder~\cite{conch}.
For each WSI, all retained tissue patches were represented as a variable-length
set of patch embeddings together with their spatial coordinates. These
precomputed patch features were kept fixed during training of the report
generation model.

To provide complementary slide-level context, we additionally extracted global
representations using TITAN~\cite{titan} and the deep slide representation from
GECKO~\cite{gecko}. The TITAN and GECKO deep features were independently
projected to the common model dimensionality and subsequently concatenated and
fused to construct the initial slide-level contextual representation used by
SCOUT.

GECKO was also used to derive case-specific concept features over the fixed
expert-curated diagnostic concept vocabulary described above. These concept
features were projected to the shared latent dimensionality before being
provided to the context-aware encoder. Detailed preprocessing procedures, including tissue-region identification, patch extraction, feature-extractor configurations, and construction of the CONCHv1.5, TITAN, and GECKO representations, are provided in Appendix~\ref{app:feature_extraction_details}.

\subsection{Concept Extraction and Representation}

% Diagnostic concepts were derived through a hybrid pipeline combining large language model-based extraction and expert validation, following the setup in~\cite{gecko}. This process ensured both scalability and clinical relevance. Concept representations were obtained using GECKO~\cite{gecko}, producing embeddings aligned with the language modeling space. These embeddings were treated as structured semantic priors and used as conditioning signals during both encoding and decoding. Additional feature extraction details are provided in Appendix~\ref{app:feature_extraction_details}, reproducibility and computational-overhead details are provided in Appendix~\ref{app:training_reproducibility}, and the curated concept sets are reported in Appendix~\ref{app:brca_gecko_concepts}, Table~\ref{tab:brca_gecko_concepts}, and Table~\ref{tab:reg_histai_concepts}.

The diagnostic concept vocabulary was constructed using the LLM-assisted and
expert-curated procedure described above. For each WSI, GECKO was used to
derive a case-specific feature vector indexed by this fixed concept
vocabulary. The concept vocabulary remained fixed across training and
inference, and no target-report text was used to construct concept features at
test time. The resulting vector was transformed by the concept adapter before
being supplied to SCOUT. Complete concept lists and extraction details are
provided in Appendix~\ref{app:feature_extraction_details},
Table~\ref{tab:brca_gecko_concepts}, and Table~\ref{tab:reg_histai_concepts}.

% Explicit diagnostic concepts were obtained by combining: structured extraction using a large language model, and manual verification through consultation with an expert pathologist.

% Each concept was embedded using GECKO\cite{gecko}, producing a fixed-dimensional representation aligned with the language model space. Concept embeddings were treated as structured conditioning signals and were not generated by the model itself.

% \subsection{Model Training}
% The model was trained to minimize the negative log-likelihood of the target report tokens:

% \begin{equation}
% \mathcal{L}_{\text{NLL}}
% =
% - \sum_{t=1}^{T}
% \log p(y_t \mid y_{<t}, \mathbf{X}),
% \end{equation}

% where X denotes the multi-modal input features.

\subsection{Implementation Details}

SCOUT was implemented in PyTorch. All models were trained using the AdamW
optimizer with weight decay applied to all non-normalization parameters
~\cite{adamw}. The learning rate was adjusted using
ReduceLROnPlateau based on validation loss.  To mitigate overfitting, dropout was applied within the attention, feed-forward, feature-adaptation, and FiLM modulation modules. Training was conducted for up to 100 epochs, with model selection based on validation performance. At inference time, reports were generated autoregressively using beam search with fixed decoding parameters across validation and test sets. All methods used the same CONCHv1.5 patch features to reduce differences attributable to the common visual input pipeline.
% Models were trained using the AdamW optimizer with weight decay applied to all non-normalization parameters. Learning rates were scheduled using cosine annealing with warm restarts to stabilize optimization without reliance on early stopping. Training was performed for a fixed number of epochs, and no test-set feedback was used during training or checkpoint selection.
% To reduce overfitting, we employed:
% \begin{itemize}
%     \item dropout in attention and feed-forward layers,
%     \item modality-specific dropout in FiLM conditioning networks,
%     \item and layer-wise aggregation to prevent over-reliance on deep representations.
% \end{itemize}
% At inference time, reports were generated autoregressively using beam search with beam width B. Temperature scaling was applied to control output entropy.
Unless otherwise specified, decoding parameters were fixed across validation and test sets to ensure comparability. 

At inference, reports are generated autoregressively from the beginning-of-sequence token. The implementation supports beam and diverse-beam decoding together with length penalties, diversity penalties, temperature scaling, and trigram blocking. Generation terminates when the end-of-sequence token is produced or the maximum sequence length is reached. Full hyperparameters, hardware specifications, and computational details are provided in Appendix~\ref{app:training_reproducibility}.

\paragraph{Baseline implementation}

We compared SCOUT with WSI-Caption~\cite{wsi_caption}, HistGen~\cite{histgen}, and Bi-Gen~\cite{bigen}. All methods used the same dataset partitions, CONCHv1.5 patch features, and
report preprocessing pipeline. Bi-Gen used the CONCH text encoder to construct its knowledge bank, whereas SCOUT additionally incorporated the TITAN and GECKO representations defined above as components of the proposed multimodal architecture. Architecture-specific hyperparameters followed the corresponding published implementations, while the fixed data partitions and common patch-feature pipeline were shared across methods.

Using a common feature-extraction and preprocessing pipeline reduces differences arising from the input representation and permits more direct comparison of the report-generation architectures.

\subsection{Evaluation Metrics}
Generated reports were evaluated using standard natural language generation metrics, which quantify complementary forms of lexical agreement with the
reference report. These metrics are:
\begin{itemize}
    \item BLEU (1--4)~\cite{bleu}, which measures n-gram overlap between generated and reference reports and is reported from unigram to four-gram precision.
    \item ROUGE-L~\cite{rouge}, which captures the longest common subsequence between the generated and reference text and reflects sentence-level content preservation.
    \item METEOR~\cite{meteor}, which evaluates unigram alignment while accounting for stemming and synonymy, providing a complementary measure of semantic similarity.
\end{itemize}

For REG 2025, we additionally evaluated structured clinical report fidelity
using the Clinical Report Quality Score (CRQS) introduced in
PathReportEval~\cite{pathreporteval}. CRQS evaluates generated and reference
reports after conversion into structured clinical attributes and combines
four components: Clinical Fact Coverage (CFC), Key Information Recall (KIR),
Hallucination Rate (HR), and Clinical Discordance Score (CDS). Higher CRQS
values indicate greater agreement with clinically relevant information in the
reference report. The complete attribute schema, extraction procedure,
component definitions, aggregation procedure, and validation of CRQS are
described in PathReportEval~\cite{pathreporteval}.
\section{Related Work}

\paragraph{Automated Medical Report Generation:}

Automated generation of clinical reports from medical images has been studied extensively in radiology, where the task is commonly framed as translating imaging evidence into structured or semi-structured diagnostic narratives\cite{jin2026grounded, gao2025s2d, che2025llm, wang2025activating, nicolson2023improving}. Early approaches focused on template-based or retrieval-based systems that mapped detected visual findings to predefined report sections. While effective in constrained settings, these systems lacked flexibility and did not generalize well to diverse pathologies or reporting styles.

The advent of deep learning shifted the field toward end-to-end image-to-text generation, drawing on encoder–decoder architectures combining convolutional neural networks (CNNs) with recurrent neural networks (RNNs), originally developed for natural-image captioning\cite{vinyals2015show}. These ideas were subsequently adapted to medical report generation, with later transformer-based approaches introducing attention mechanism to align image regions with textual tokens\cite{sengupta2024automatic,chen2020generating,chen2022cross}.

% The advent of deep learning shifted the field toward end-to-end image-to-text generation, inspired by image captioning in natural images. Encoder–decoder architectures combining convolutional neural networks (CNNs) with recurrent neural networks (RNNs)\cite{vinyals2015show} were among the first to demonstrate feasibility for radiology report generation. Subsequent works introduced attention mechanisms to align image regions with textual tokens\cite{sengupta2024automatic}, significantly improving descriptive accuracy and coherence.

% The development of end-to-end image-to-text models shifted the field toward neural encoder--decoder architectures. Initial approaches combined convolutional image encoders with recurrent language models
% \cite{vinyals2015show}, followed by attention-based methods that aligned image regions with generated words or sentences
% \cite{sengupta2024automatic}.

% More recent systems use Transformer-based architectures, hierarchical decoders, contrastive objectives, structured disease representations, and clinically motivated reward functions to improve factual consistency and reduce hallucination
% \cite{chen2020generating,chen2022cross,nicolson2023improving}.

More recently, transformer-based architectures have become the dominant paradigm. Models such as Show-Attend-and-Tell variants, hierarchical LSTM–Transformer hybrids, and fully transformer-based frameworks have been proposed for chest X-ray and CT report generation\cite{chen2020generating,chen2022cross,nicolson2023improving}. Several studies emphasized hierarchical decoding, where sentence-level and word-level generators reflect the structured nature of radiology reports. Others incorporated graph-based disease dependencies, reinforcement learning with clinical reward functions, or contrastive learning objectives to improve factual correctness and reduce hallucinations.
% Despite these advances, radiology report generation typically benefits from relatively low-resolution images, standardized acquisition protocols, and well-defined diagnostic vocabularies, assumptions that do not hold in histopathology.

These advances established many of the architectural principles used in medical report generation. However, radiology models generally operate on images with substantially lower resolution and fewer spatial instances than whole-slide pathology images. Histopathology additionally requires reasoning across cellular, regional, and slide-level scales, making direct adaptation of conventional image-to-report architectures difficult.

% \paragraph{Vision-language models in medical imaging.}
% More broadly, vision-language models have shown strong performance in medical report generation tasks, including radiology and pathology. These models commonly adopt transformer-based architectures with cross-attention between visual and textual representations. However, most approaches treat visual features as static inputs and do not explicitly model the hierarchical and multi-scale nature of pathology data, limiting their ability to capture complex spatial and semantic relationships.

\paragraph{Vision–Language Models in Medical Imaging:}
Large-scale vision-language models such as CLIP and ALIGN demonstrated that contrastive pretraining on paired images and text can produce transferable cross-modal representations. These models inspired a wave of multimodal architectures for medical applications, including contrastive pretraining on image–text pairs and prompt-based adaptation. 

Most general-purpose vision--language models, however, represent an image using a small number of global tokens or features. Such formulations are poorly suited to gigapixel WSIs, which may contain thousands of spatially distributed regions and substantial morphological heterogeneity. Moreover, although these models learn semantic alignment between images and text, they do not inherently encode domain-specific diagnostic concepts or specify how representations should be organized to support clinical interpretation. Contrastive alignment alone therefore does not determine how local tissue morphology, whole-slide architecture, and diagnostic concepts should interact during report generation.

\paragraph{Concept-guided and interpretable learning in pathology:}
Concept-based learning introduces intermediate semantic representations intended to connect model predictions with human-understandable attributes. In computational pathology, these concepts may describe cellular morphology, tissue architecture, tumor grade, stromal characteristics, or other diagnostically relevant findings. 

Recent approaches such as SI-MIL~\cite{si_mil} and GECKO~\cite{gecko} associate learned visual representations with clinically meaningful concepts derived from expert knowledge, textual descriptions, or large language models. These methods can support concept discovery, concept-based classification, post hoc analysis of model predictions. They therefore provide a useful mechanism for organizing visual representations around clinically relevant semantic attributes. Most existing concept-based pathology methods are designed for classification or representation analysis rather than free-text generation. Concept information is commonly used as an auxiliary supervision signal, a bottleneck representation, or an explanatory layer after prediction. In contrast, relatively little work has studied how explicit diagnostic concepts can act as conditioning signals that continuously influence visual representation learning and subsequent language generation.
% While these approaches provide useful insights, they have not been adopted for report generation problems previously to the best of our knowledge. No similar concept based signals have been used in .
% \paragraph{Concept-based and interpretable models.}
% Concept-based learning approaches aim to improve interpretability by introducing intermediate semantic representations. 

% In medical imaging, such methods often rely on predefined concepts or post hoc explanations to link predictions to human-interpretable features. While these approaches provide useful insights, they are typically not integrated directly into the generative process, and therefore do not influence the formation of textual outputs in a structured manner.

\paragraph{Whole-slide Pathology Report Generation}
Computational pathology introduces distinct challenges for report generation because WSIs are gigapixel images with hierarchical and spatially heterogeneous tissue structure. Early deep learning approaches primarily addressed patch-level classification or slide-level prediction using multiple instance learning (MIL). These methods aggregate features from sampled tissue regions to produce slide-level labels, but they do not generate detailed textual descriptions of the underlying findings.

Extending such models to free-text report generation is non-trivial. Diagnostic pathology reports summarize local cytologic features, regional architectural patterns, tumor extent, grade, margins, and other high-level diagnostic abstractions. A successful model must therefore compress evidence from many spatial instances while preserving the relationships between local observations and slide-level conclusions.

WSI-Caption~\cite{wsi_caption} formulated WSI-to-report generation as a multiple-instance sequence-generation problem. The method aggregates patch-level representations before decoding a slide-level report, demonstrating that report generation is feasible at WSI scale. Additionally, WSI-Caption showed that language supervision can improve the utility of learned slide representations for downstream diagnostic tasks. However, the visual features remain fixed after extraction, and the model does not explicitly represent diagnostic concepts or model separate reasoning pathways for local, global, or semantic information.

HistGen~\cite{histgen} addressed the mismatch between dense visual evidence and concise pathology reports through a local--global hierarchical encoder and a cross-modal context module. Its hierarchical architecture models visual information at multiple granularities and improves alignment between visual representations and linguistic structure. This represents an important advance in structured WSI representation learning. However, HistGen primarily emphasizes hierarchical aggregation and cross-modal alignment rather than adaptive conditioning or explicit concept-level reasoning during generation. Its contextual representations therefore do not progressively modulate local patch features using explicit diagnostic concepts across network depth.

% Bi-Gen~\cite{bigen} introduced a knowledge-enhanced framework that retrieves semantically relevant historical information from an external knowledge bank associated with visually salient regions. Retrieved knowledge and WSI representations are processed using a bi-modal learning strategy with shared cross-attention, enriching the visual representation with clinically relevant textual context. This design improves semantic consistency but depends on external retrieval and integrates knowledge primarily through cross-modal tokens and downstream fusion. Consequently, the interaction between retrieved knowledge and local visual evidence remains indirect, and the method does not explicitly separate iterative representation refinement from downstream multimodal fusion or expose token-level modality contributions during generation.

Bi-Gen~\cite{bigen} further introduced a knowledge-enhanced framework that retrieves semantically relevant historical information from an external knowledge bank associated with visually salient regions. Retrieved knowledge and WSI representations are processed using a bi-modal learning strategy with shared cross-attention, enriching the visual representation with clinically relevant textual context. This design improves semantic consistency but depends on external retrieval and integrates knowledge primarily through cross-modal tokens and downstream fusion. Consequently, the interaction between retrieved knowledge and local visual evidence remains indirect, and the method does not explicitly separate iterative representation refinement from downstream multimodal fusion or provide transparent control over how different modalities influence individual report statements.

Together, these approaches have substantially advanced WSI report generation through multiple-instance modeling, hierarchical aggregation, and knowledge augmentation. However, they generally rely on visual features that remain fixed after extraction or introduce semantic context after the principal visual representations have already been formed. Auxiliary modalities are also often derived from the same patch-level features or incorporated primarily through aggregation and downstream fusion. These designs therefore provide limited mechanisms for allowing global tissue context or explicit diagnostic concepts to progressively alter the interpretation of local morphology throughout the representation hierarchy.

SCOUT differs from these approaches in two principal respects. First, it represents patch-level morphology, whole-slide context, and explicit histopathologic concepts as distinct but interacting information sources and iteratively updates them across encoder layers. Diagnostic concepts therefore act as active conditioning signals rather than solely as auxiliary supervision or retrieved context. Second, the decoder attends separately to the three representation streams and uses learned token-level gates to adapt their relative contributions during report generation. This formulation enables progressive context refinement during encoding and provides inspectable estimates of modality use during decoding.

\section{Discussions and Conclusion}

In this work, we introduced SCOUT, a multimodal framework for pathology report
generation that integrates local patch-level morphology, global slide context,
and structured diagnostic concepts through iterative contextual updating and
adaptive multimodal fusion. Across TCGA-BRCA, HistAI, and REG-2025, SCOUT
consistently improved report-generation performance relative to recent
whole-slide report generation methods across most lexical metrics, while also
improving structured clinical report fidelity on REG-2025 as measured by CRQS.
Together with the ablation experiments, these results support the value of
combining complementary visual and semantic representations with adaptive
fusion rather than relying on a single representation stream or static feature
integration.

A central methodological component of SCOUT is its context-aware encoder. The
model maintains an evolving patch backbone together with slide-conditioned and
concept-conditioned feature streams. At each encoder depth, the current patch
representation is used to construct the two context-conditioned streams, which
are subsequently pooled to update the corresponding slide and concept states
used at the following layer. Contextual information therefore evolves together
with increasingly transformed visual representations rather than remaining
fixed throughout the encoder. Learned depth-wise aggregation further allows
the final patch, slide-conditioned, and concept-conditioned representations to
draw on different encoder depths. The ablation experiments support a
contribution from this iterative contextual formulation, although the effect
was not uniform across all lexical metrics, indicating that contextual
modulation provides complementary rather than universally monotonic gains.

The decoder provides a second level of multimodal integration. Rather than
combining the three encoder outputs before language generation, SCOUT preserves
separate cross-attention pathways for the patch, slide-conditioned, and
concept-conditioned streams and combines their outputs using feature-wise
multi-head gated fusion. This mechanism allows the relative allocation to each
representation stream to vary across generated tokens, fusion heads, and
within-head feature dimensions. Gated fusion produced the strongest
performance in the ablation study, suggesting that adaptive combination of
heterogeneous representations is preferable to the static fusion strategies
examined here.

Analysis of the learned gates further showed that the decoder did not collapse
onto a single representation stream. Patch-derived information received the
largest average contribution across the REG-2025 test cohort, while
concept-conditioned and slide-conditioned representations retained substantial
non-zero contributions. Case-level cross-attention visualizations also showed
distinct spatial patterns across the three streams. These analyses provide an
inspectable view of how the model allocates information internally during
generation, but they should not be interpreted as causal explanations of
individual predictions. Establishing causal faithfulness of these signals
would require targeted intervention or perturbation experiments beyond the
scope of the present study.

The explicit use of diagnostic concepts provides a structured mechanism for
incorporating domain knowledge into report generation. In SCOUT, concepts are
not used only as auxiliary supervision or as post hoc annotations; instead,
case-specific GECKO concept features derived over a fixed expert-curated
vocabulary form one of the contextual inputs to the encoder and one of the
representation streams available to the decoder. Importantly, the concept
vocabulary is fixed before test-time evaluation and no target-report text is
used to construct concept features at inference. This design provides a
practical route for incorporating pathology-specific semantic structure while
retaining separation between the concept-conditioned and purely visual
representations.

The consistent improvements observed across TCGA-BRCA, HistAI, and REG-2025
are notable because these datasets differ substantially in cohort size,
disease composition, and reporting characteristics. These experiments do not
constitute a cross-dataset transfer evaluation, because models were trained and
evaluated within each dataset. They therefore support consistency of the
method across heterogeneous report-generation settings rather than
generalization under domain shift. Evaluating SCOUT under explicit
cross-institutional or cross-dataset transfer will be an important direction
for future work.

The CRQS analysis complements the conventional language-generation metrics by
examining structured clinical report content. On REG-2025, the improvement in
CRQS indicates that the lexical gains obtained by SCOUT were accompanied by
better preservation of clinically relevant report attributes. This distinction
is important because n-gram and sequence-overlap metrics alone cannot
distinguish fluent reports from reports containing clinically consequential
omissions, unsupported findings, or discordant attributes. Nevertheless, CRQS
remains an automated structured evaluation metric and should not be considered
a substitute for direct assessment by practicing pathologists.

Several limitations should therefore be considered. First, the present study
does not include a blinded evaluation of generated reports by independent
pathologists. Although CRQS provides structured assessment of clinically
relevant report attributes, future studies should examine agreement between
automated metrics and expert judgments of diagnostic correctness,
completeness, and clinical utility. Second, the diagnostic concept vocabulary
depends on an LLM-assisted and expert-curated construction process. Errors,
omissions, or differences in concept coverage may affect the resulting GECKO
features and consequently model behavior. Developing scalable procedures for
constructing and validating concept vocabularies across organs and disease
types remains an important area for further investigation. Third, the modality
gate and cross-attention analyses are descriptive model-internal signals rather
than causal explanations. Intervention-based analyses will be required to
determine whether changes in these signals faithfully correspond to changes in
specific generated findings. Fourth, our evaluation assesses performance
within each dataset and does not directly measure robustness to cross-dataset,
cross-institutional, or acquisition-domain shifts. Finally, SCOUT relies on
multiple pretrained pathology encoders and additional contextual and fusion
modules, increasing computational complexity relative to single-stream report
generation architectures, although feature extraction is performed offline and
the additional train-time computation is confined to the lightweight
adaptation, contextual encoding, and fusion modules.

Future work could address these limitations in several directions. Joint or
parameter-efficient adaptation of the pretrained visual encoders may allow
feature extraction to become more task-specific while retaining the advantages
of foundation-model initialization. Concept representations could also be
extended beyond fixed vocabularies using dynamically retrieved or hierarchically
organized pathology knowledge. Prospective expert evaluation and explicit
cross-domain transfer experiments would provide a stronger assessment of
clinical reliability and robustness. In addition, token- or finding-specific
intervention studies could clarify how different representation streams affect
individual report statements. More broadly, the separation of evolving visual
representations, structured domain context, and adaptive multimodal fusion may
be applicable to other medical image-to-text tasks in which heterogeneous
visual evidence must be translated into structured clinical narratives.

In conclusion, SCOUT provides a framework for pathology report generation that
combines an evolving patch representation with recursively updated
slide-conditioned and concept-conditioned representations and adaptively fuses
these information streams during language generation. The method achieved
consistent improvements across three heterogeneous pathology report datasets
and improved structured clinical report fidelity on REG-2025. These findings
support iterative contextual representation updating and adaptive multimodal
fusion as useful design principles for incorporating structured domain
knowledge into pathology vision--language models, while highlighting the need
for future expert validation and cross-domain evaluation before drawing
conclusions about clinical utility or generalization.

\section*{Acknowledgments}
This research was partially supported
by NIH grants 1R01CA297843-01, 1R03DE033489-01A1,
and NSF grant 2442053. The content is solely the responsibility of the authors and does not necessarily represent the
official views of the NIH.

\section*{Data availability}
The datasets analyzed in this study are available from their respective original sources: TCGA-BRCA through The Cancer Genome Atlas and PathText\cite{wsi_caption}, HistAI through its published repository\cite{histai_dataset}, and REG-2025 through the MICCAI 2025 REG challenge data-access mechanism\cite{reg2025_challenge}. The fixed training, validation, and test split definitions used for all experiments are available in the project repository at \url{https://github.com/surykntsingh/SCOUT}.

\section*{Code availability}
Source code for SCOUT, including training and evaluation scripts, experimental configuration files, and the fixed dataset split definitions used in this study, is available at \url{https://github.com/surykntsingh/SCOUT}.

\section*{Competing interests}

The authors declare no competing interests.

\section*{Author contributions}

S.S. conceptualized the project, developed the methodology, implemented the models, conducted the experiments, analyzed the results, and drafted the manuscript. S.K. contributed to preparing visual features and provided methodological input. J.S. contributed to interpretation of the results, clinical evaluation,  and manuscript revision. P.P. supervised the project, contributed to project design, interpretation of the results, and revised the manuscript. All authors reviewed and approved the final manuscript.

%%%% References %%%%%%%%%%%%%%%%%%%%%%%%%%%%%%%%%%%%%%%%%%%%

\bibliographystyle{abbrv}
\bibliography{references}

%%%% Appendix %%%%%%%%%%%%%%%%%%%%%%%%%%%%%%%%%%%%%%%%%%%%%%

\clearpage
\appendix

\section{Appendix}

\subsection{Additional Architecture Details}
\label{app:architecture}

This section provides implementation details omitted from the main manuscript due to space constraints. We describe the additional details about the architecture, modality adaptation pipeline, iterative encoder refinement mechanism, encoder/decoder design, and implementation choices necessary for reproducibility.

\subsubsection{Overview Recap}
Given a whole-slide image, patch-level features are extracted using CONCHv1.5, slide-level representations are obtained using TITAN, and concept-level features are generated using GECKO. Since these representations originate from independently pretrained foundation models, each modality is first adapted to a common latent space using modality-specific projection networks before multimodal interaction.

% Unlike conventional report generation models that directly fuse heterogeneous features, SCOUT separates representation learning into two stages. The encoder progressively refines visual representations through iterative contextual conditioning, while the decoder performs token-level multimodal reasoning using adaptive gated fusion. This separation allows contextual information to influence feature learning before language generation begins.
Unlike approaches that immediately collapse heterogeneous inputs into a
single representation, SCOUT maintains an evolving patch backbone together
with slide-conditioned and concept-conditioned spatial streams. At each
encoder depth, the current patch representation is used to construct the two
context-conditioned streams, which are subsequently pooled to update their
corresponding contextual states. The updated contextual states condition the
next encoder depth, while the patch backbone continues through successive
Transformer and PAM operations.

%%%%%%%%%%%%%%%%%%%%%%%%%%%%%%%%%%%%%%%%%%%%%%%%%%%%%%%%%%%%%

\subsubsection{Encoder details}
\label{app:encoder_details}
\begin{figure}[t]
\centering
\includegraphics[width=0.98\textwidth]{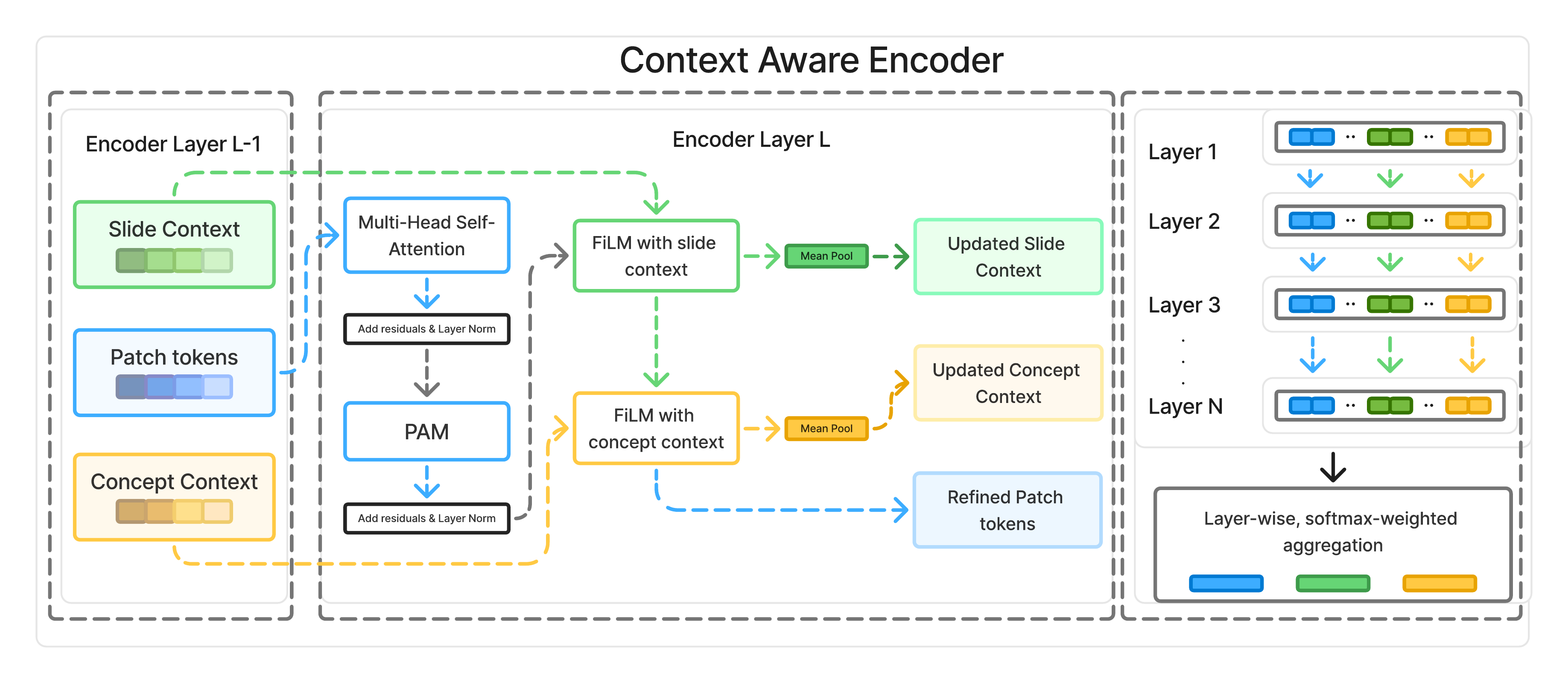}
\caption{
\textbf{Detailed view of the context-aware encoder.}
The encoder refines patch-level representations through multi-head self-attention followed by a Position Aware Module (PAM), which captures local spatial interactions across multiple receptive fields. At each encoder layer $\ell$, the updated patch tokens are conditioned using slide-level context $s^{(\ell-1)}$ and concept-level context $c^{(\ell-1)}$ through FiLM modulation. The resulting slide- and concept-conditioned patch representations are then pooled to update the contextual states $s^{(\ell)}$ and $c^{(\ell)}$, enabling iterative co-evolution of local morphology, global slide context, and diagnostic concepts across encoder depth. Final patch, slide, and concept features are obtained through learnable softmax-weighted aggregation over all encoder layers.
}
\label{fig:encoder_details}
\end{figure}

Figure~\ref{fig:encoder_details} provides an implementation-level view of the context-aware encoder. While the main manuscript describes the encoder mathematically, this section clarifies the practical data flow through each encoder layer and highlights the components that support iterative contextual refinement. Each encoder layer receives patch tokens, a slide-level context vector, and a concept-level context representation. The patch tokens are first processed by standard multi-head self-attention with residual connections and normalization. The resulting patch sequence is then passed through the Position-Aware Module (PAM), which introduces a local spatial inductive bias by reshaping the patch sequence into a two-dimensional patch grid and applying depth-wise convolutional filters with multiple receptive fields. In our implementation, PAM uses $13 \times 13$, $7 \times 7$, and $3 \times 3$ depth-wise convolution branches, whose outputs are combined with the original patch features through residual addition. This allows the encoder to preserve global token interactions from self-attention while explicitly modeling local neighborhood structure among nearby WSI patches. 

After PAM, the patch representations are separately modulated by slide-level and concept-level context using FiLM conditioning. The slide-conditioned and concept-conditioned patch representations are then pooled across the patch dimension to update the slide and concept states for the next encoder layer. Therefore, the contextual inputs to deeper layers are not fixed copies of the original slide and concept features; instead, they are progressively refined based on the evolving visual representation. This creates an iterative co-evolution process in which the evolving patch representation progressively updates the global and semantic contextual states, which subsequently condition the corresponding slide- and concept-conditioned branches at the next encoder depth. The encoder stores patch-, slide-, and concept-conditioned representations from all layers. Rather than relying only on the final encoder layer, SCOUT learns separate softmax-normalized layer aggregation weights for the patch, slide, and concept streams. This allows each stream to emphasize the encoder depths most useful for downstream report generation. The resulting aggregated patch, slide-conditioned, and concept-conditioned features are passed to the multimodal decoder, where they are used as separate memory sources for cross-attention and adaptive gated fusion.

%%%%%%%%%%%%%%%%%%%%%%%%%%%%%%%%%%%%%%%%%%%%%%%%%%%%%%%%%%%%%

\subsubsection{Input Representation}

For each whole-slide image, three complementary feature streams are extracted.

\begin{itemize}
    \item \textbf{Patch features} $\mathbf{P}\in\mathbb{R}^{N\times d_p}$ are extracted using the CONCHv1.5 pathology foundation model, where $N$ denotes the number of image patches.
    
    \item \textbf{Slide features} $\mathbf{s}\in\mathbb{R}^{d_t}$ are extracted using TITAN and summarize global tissue architecture.
    
    \item \textbf{Concept features} $\mathbf{q}\in\mathbb{R}^{K}$ are obtained from GECKO, where $K$ denotes the size of the curated diagnostic-concept vocabulary.
\end{itemize}

All three modalities are projected into a shared latent dimension $d$ before entering the Transformer.

%%%%%%%%%%%%%%%%%%%%%%%%%%%%%%%%%%%%%%%%%%%%%%%%%%%%%%%%%%%%%

\subsubsection{Modality Adaptation Networks}

Although all modalities are represented as feature embeddings, they originate from independently pretrained foundation models and therefore exhibit substantially different statistical distributions and embedding geometries. Rather than using a single linear projection, SCOUT employs modality-specific multilayer perceptrons for feature adaptation.

Each adapter consists of

\begin{center}
Linear $\rightarrow$ ReLU $\rightarrow$
Linear $\rightarrow$ ReLU $\rightarrow$
Dropout $\rightarrow$
Linear $\rightarrow$
LayerNorm.
\end{center}

Separate adapters are used for patch, slide, GECKO deep, and concept embeddings. This design allows each modality to preserve its semantic structure while adapting to the downstream report generation task.

%%%%%%%%%%%%%%%%%%%%%%%%%%%%%%%%%%%%%%%%%%%%%%%%%%%%%%%%%%%%%

\subsubsection{Learnable Prompt Token}

Before entering the encoder, a learnable prompt token is prepended to the sequence of adapted patch embeddings. This token is randomly initialized and optimized jointly with the remainder of the network.

Unlike CLS tokens used for image classification, the prompt token serves as a global latent representation that facilitates information exchange across image patches during self-attention. It additionally provides a stable reference representation for subsequent contextual conditioning throughout the encoder.

%%%%%%%%%%%%%%%%%%%%%%%%%%%%%%%%%%%%%%%%%%%%%%%%%%%%%%%%%%%%%

\subsubsection{Construction of Slide-Level Context}
\label{app:slide_features_construction}
The slide representation used for contextual conditioning is constructed from two complementary sources.

First, TITAN provides a global whole-slide representation summarizing tissue architecture. Second, GECKO deep embeddings capture high-level semantic information learned during pathology image--text pretraining. These two representations are concatenated and passed through an additional projection network to obtain the initial slide context used by the encoder.

This design allows global contextual information to incorporate both morphological organization and semantic priors before iterative refinement begins.

%%%%%%%%%%%%%%%%%%%%%%%%%%%%%%%%%%%%%%%%%%%%%%%%%%%%%%%%%%%%%

\subsubsection{Iterative Co-evolution of Visual and Semantic Representations}

A distinguishing characteristic of SCOUT is that contextual representations are not treated as fixed conditioning signals. Instead, these contextual states recursively co-evolve across encoder depth.

At encoder layer $\ell$, the patch representation is first updated using
multi-head self-attention followed by the Position-Aware Module (PAM),
producing $\mathbf{P}^{(\ell)}$. The resulting patch representation is then
independently conditioned by the slide and concept states from the preceding
layer using separate FiLM modules, producing the spatially resolved
slide-conditioned and concept-conditioned representations
$\mathbf{S}^{(\ell)}$ and $\mathbf{C}^{(\ell)}$, respectively.

The corresponding contextual states are subsequently updated by pooling these
conditioned representations over the visual-token dimension:

\begin{equation}
\mathbf{s}^{(\ell)}
=
\operatorname{MeanPool}\left(
\mathbf{S}^{(\ell)}
\right),
\end{equation}

\begin{equation}
\mathbf{c}^{(\ell)}
=
\operatorname{MeanPool}\left(
\mathbf{C}^{(\ell)}
\right).
\end{equation}

The updated states $\mathbf{s}^{(\ell)}$ and $\mathbf{c}^{(\ell)}$ then serve
as conditioning signals for the slide- and concept-conditioned FiLM branches
at encoder layer $\ell+1$. Thus, contextual information is progressively
updated on the basis of increasingly transformed visual representations rather
than remaining fixed throughout the encoder. Importantly, the patch backbone
itself continues through successive Transformer and PAM operations, while the
slide- and concept-conditioned branches are reconstructed from the current
patch representation at each depth.

We refer to this recursive evolution of the visual representation together
with its associated contextual states as \emph{iterative co-evolution}.
This formulation preserves a distinct patch backbone while allowing the
contextual representations used for conditioning to adapt progressively
across encoder depth.

%%%%%%%%%%%%%%%%%%%%%%%%%%%%%%%%%%%%%%%%%%%%%%%%%%%%%%%%%%%%%

\subsubsection{Position Aware Module}
\label{app:pam}
Following each self-attention layer, patch representations are processed by the Position-Aware Module (PAM) proposed by P. Chen et al.~\cite{wsi_caption}.

Instead of applying additional Transformer attention, PAM reshapes the patch sequence into its two-dimensional spatial layout and applies depth-wise convolutions with kernel sizes

\[
13\times13,\qquad
7\times7,\qquad
3\times3.
\]

Outputs from all convolutional branches are combined through residual addition before being flattened back into the token sequence.

The multiple receptive fields allow the encoder to capture local glandular morphology together with larger architectural patterns that are difficult to model using self-attention alone.

%%%%%%%%%%%%%%%%%%%%%%%%%%%%%%%%%%%%%%%%%%%%%%%%%%%%%%%%%%%%%

\subsubsection{Layer-wise Feature Aggregation}

Rather than using only the final encoder layer, SCOUT aggregates representations from all encoder layers.

For each modality independently, learnable layer weights are normalized using a softmax,

\begin{equation}
w_\ell
=
\frac{\exp(\theta_\ell)}
{\sum_k \exp(\theta_k)},
\end{equation}

and the final representation is computed as

\begin{equation}
\mathbf{F}
=
\sum_{\ell=1}^{L}
w_\ell
\mathbf{H}^{(\ell)}.
\end{equation}

Separate aggregation weights are learned for patch, slide, and concept representations, allowing different modalities to emphasize different abstraction levels.

%%%%%%%%%%%%%%%%%%%%%%%%%%%%%%%%%%%%%%%%%%%%%%%%%%%%%%%%%%%%%

\subsubsection{Decoder Architecture}
\label{app:decoder_details}
\begin{figure}[t]
\centering
\includegraphics[width=0.98\textwidth]{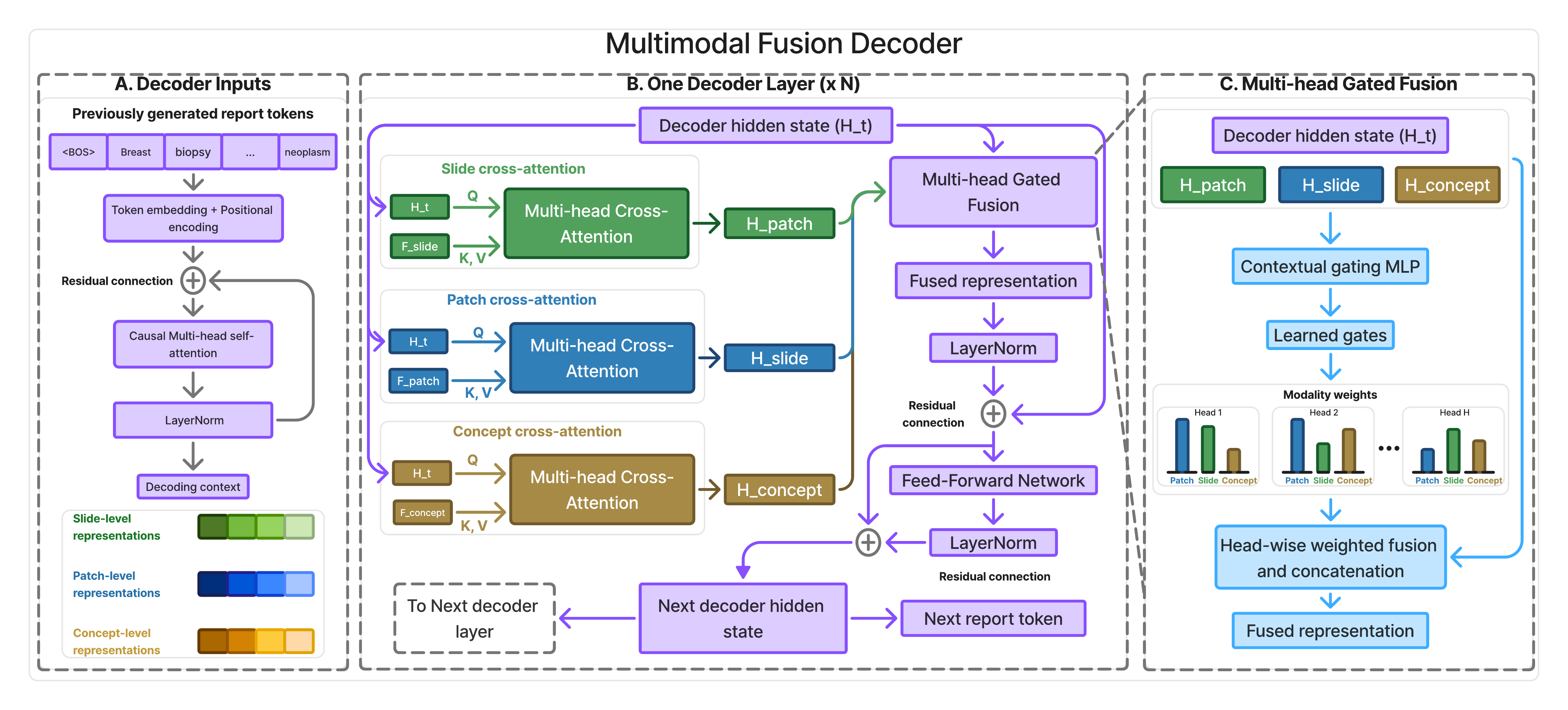}
\caption{
\textbf{Detailed view of the adaptive multimodal decoder.}
At each decoding layer, autoregressive token representations first undergo causal self-attention and then attend independently to the encoder-derived patch, slide, and concept feature streams. The resulting modality-specific attended features are projected and combined using a multi-head 3-way gated fusion module, which learns token-, head-, and feature-specific weights over local visual evidence, global slide context, and diagnostic concept information. The fused representation is passed through residual refinement and a feed-forward network before producing vocabulary logits for report generation.
}
\label{fig:decoder_details}
\end{figure}

Figure~\ref{fig:decoder_details} provides an implementation-level view of the SCOUT decoder. The decoder preserves separate cross-attention pathways for patch, slide, and concept representations before combining them with adaptive multimodal gated fusion, enabling token-level selection of complementary evidence during report generation.

Each decoder layer consists of six sequential operations:

\begin{enumerate}
\item causal self-attention,
\item cross-attention over patch representations,
\item cross-attention over slide representations,
\item cross-attention over concept representations,
\item adaptive multimodal gated fusion,
\item position-wise feed-forward network.
\end{enumerate}

Maintaining independent cross-attention pathways preserves modality-specific information until the final fusion stage.

%%%%%%%%%%%%%%%%%%%%%%%%%%%%%%%%%%%%%%%%%%%%%%%%%%%%%%%%%%%%%

\subsubsection{Multi-head Gated Fusion}

SCOUT combines the patch-, slide-conditioned, and concept-conditioned decoder representations using a feature-wise multi-head gating mechanism. Unlike static concatenation, the fusion weights are conditioned on the current decoder state and can vary across generated tokens and feature dimensions.

At decoder layer $j$, let
$\widetilde{\mathbf{D}}{p}$,
$\widetilde{\mathbf{D}}{s}$, and
$\widetilde{\mathbf{D}}_{c}
\in\mathbb{R}^{B\times T\times d}$
denote the representations obtained after cross-attention to the patch,
slide-conditioned, and concept-conditioned encoder streams, respectively.
Here, $B$ is the batch size, $T$ is the current decoder sequence length, and
$d$ is the model dimensionality. The three representations are first
independently projected:

$$
\mathbf{R}{p}=\mathbf{W}{p}\widetilde{\mathbf{D}}{p},
\qquad
\mathbf{R}{s}=\mathbf{W}{s}\widetilde{\mathbf{D}}{s},
\qquad
\mathbf{R}{c}=\mathbf{W}{c}\widetilde{\mathbf{D}}_{c}.
$$
To determine how these representations should be combined, the current decoder
state $\mathbf{D}$ is concatenated with all three cross-attended
representations,$$
[\mathbf{D};
\widetilde{\mathbf{D}}{p};
\widetilde{\mathbf{D}}{s};
\widetilde{\mathbf{D}}_{c}]$$ to form the gating context which is passed through the contextual gating network $\phi(\cdot)$ to produce modality-specific gating logits $\mathbf{Z}=\phi(\mathbf{Q})$. For $H$ fusion heads, each with dimensionality $d_h=d/H$, the gating logits are reshaped as $
\mathbf{Z}
\in
\mathbb{R}^{B\times T\times H\times 3\times d_h}$.

The dimensions of this tensor have the following interpretation:

\begin{itemize}
\item $t\in{1,\ldots,T}$ indexes the decoder token position;
\item $h\in{1,\ldots,H}$ indexes the fusion head;
\item $m\in{p,s,c}$ indexes the patch, slide-conditioned, or
concept-conditioned representation stream; and
\item $r\in{1,\ldots,d_h}$ indexes an individual feature dimension
within fusion head $h$.
\end{itemize}

Thus, $z_{t,h,m,r}$ is the unnormalized gating score assigned to representation
stream $m$ for feature dimension $r$ of fusion head $h$ while generating
token $t$.

The three modality scores corresponding to a fixed $(t,h,r)$ are normalized
with a temperature-scaled softmax $
\frac{
\exp\left(z_{t,h,m,r}/\tau\right)
}{
\sum_{m'}
\exp\left(z_{t,h,m',r}/\tau\right)
},
$ where $\tau$ is a learnable temperature parameter initialized to 1. Consequently $\sum_m \alpha_{t,h,m,r}=1$ for every token $t$, fusion head $h$, and within-head feature dimension $r$. Importantly, the gate is therefore \emph{feature-wise}: SCOUT does not assign
a single patch, slide, and concept weight to an entire fusion head. Instead,
each feature dimension within that head can use a different mixture of the
three representation streams. The projected representations are reshaped into the same $H$-head structure.
For token $t$, head $h$, and within-head feature dimension $r$, the fused
feature is

$$F_h =
\alpha_{t,h,p,r}R_{p,t,h,r}
+
\alpha_{t,h,s,r}R_{s,t,h,r}
+
\alpha_{t,h,c,r}R_{c,t,h,r}.
$$

In other words, each fused feature is a learned weighted combination of the
corresponding patch-, slide-conditioned, and concept-conditioned features.
The fused heads are then concatenated and projected back to the model
dimensionality $d$. The fusion module applies residual addition and LayerNorm,
followed by a feed-forward transformation, and is itself contained within a
residual decoder sublayer.

The resulting gate values provide model-internal estimates of how the decoder
allocates representation capacity among the three streams during generation.
For the cohort-level analyses in the main manuscript, these feature-wise gates
are subsequently averaged across generated tokens, fusion heads, and
within-head feature dimensions to obtain a single case-level contribution for
each representation stream. These quantities are descriptive measures of
model-internal modality allocation and are not interpreted as causal measures
of feature importance.

%%%%%%%%%%%%%%%%%%%%%%%%%%%%%%%%%%%%%%%%%%%%%%%%%%%%%%%%%%%%%

\subsubsection{Attention Maps and modality contribution}

To facilitate analysis of how the decoder accesses and combines the three
representation streams, SCOUT returns modality-specific cross-attention maps and
fusion weights from every decoder layer, including

\begin{itemize}
\item patch cross-attention,
\item slide cross-attention,
\item concept cross-attention,
\item modality fusion weights.
\end{itemize}

These outputs describe two complementary aspects of the decoding process. The
cross-attention maps indicate which spatial regions are emphasized when the
generated tokens attend to patch-, slide-, and concept-conditioned
representations. The fusion weights instead quantify the relative allocation of
the decoder to the three attended streams when forming the fused representation.
Thus, the attention maps characterize where each representation stream focuses,
whereas the gates characterize how strongly the streams are combined during
generation.

The main-text analyses use these outputs at two levels. First, mean gate weights
are summarized across the REG 2025 test cohort in
Fig.~\ref{fig:gate_distribution} to determine whether the learned fusion
mechanism collapses onto a single modality. Second, the modality-specific
attention maps are shown together with the corresponding mean gate contributions
for an individual case in Fig.~\ref{fig:attention_gate_case}. This joint
visualization connects the spatial emphasis of each stream with its relative
contribution to the generated report. The same analysis is repeated for three
additional cases in Figs.~\ref{fig:attention_gate_case_03006}--\ref{fig:attention_gate_case_09073}, providing further examples of case-dependent attention patterns while retaining the consistent ordering of modality-level gate contributions observed in the cohort analysis.

These quantities provide inspectable model-internal signals rather than causal
explanations. High attention or gate weight does not by itself establish that a
region or modality caused a particular diagnostic statement, and the case-level
figures should therefore be interpreted as qualitative complements to the
cohort-level gate analysis and quantitative report evaluation.

\subsubsection{Additional case-level attention and gated-fusion examples}
\label{app:additional_attention_cases}

\begin{figure*}[p]
    \centering
    \includegraphics[width=\linewidth]{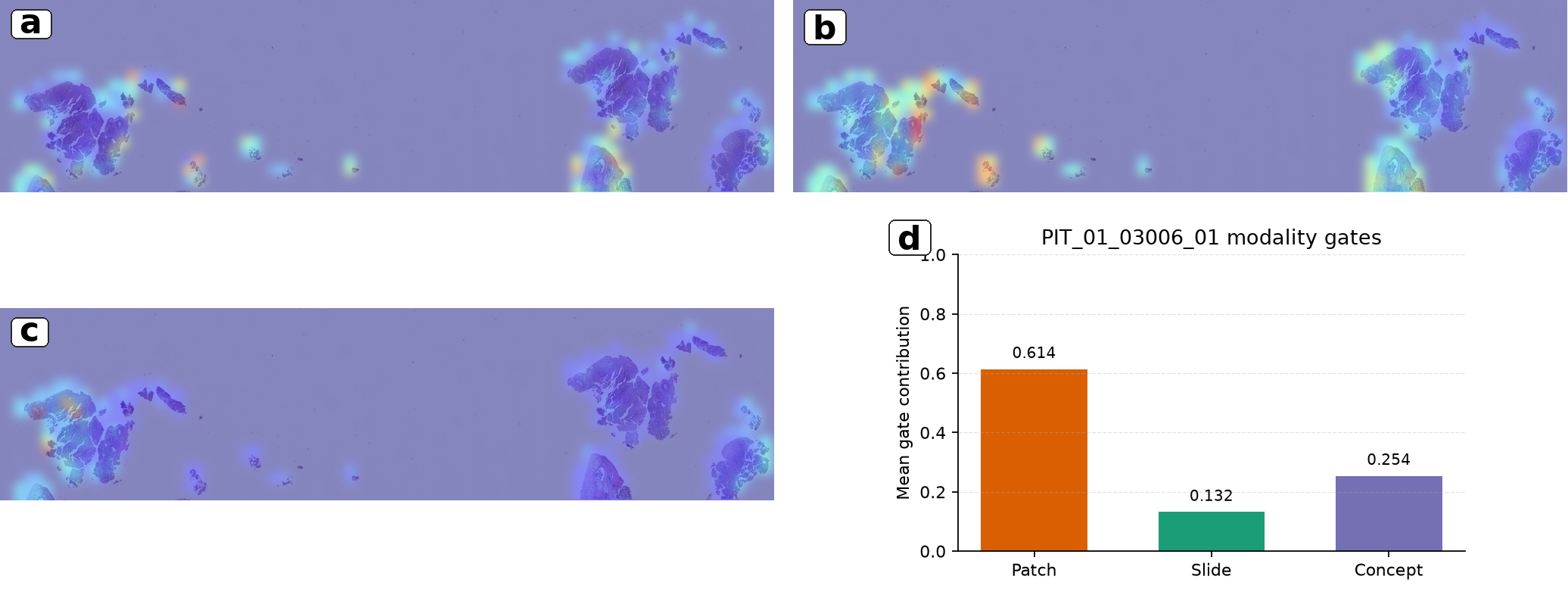}
    \caption{\textbf{Additional case-level modality analysis for case PIT\_01\_03006\_01.}
    \textbf{a--c}, Patch-, slide-, and concept-level attention maps, respectively.
    \textbf{d}, Mean decoder gate contributions for the same case. The panels
    illustrate the distinct spatial emphasis of the three representation streams
    and their relative use during report generation.}
    \label{fig:attention_gate_case_03006}
\end{figure*}

\begin{figure*}[p]
    \centering
    \includegraphics[width=\linewidth]{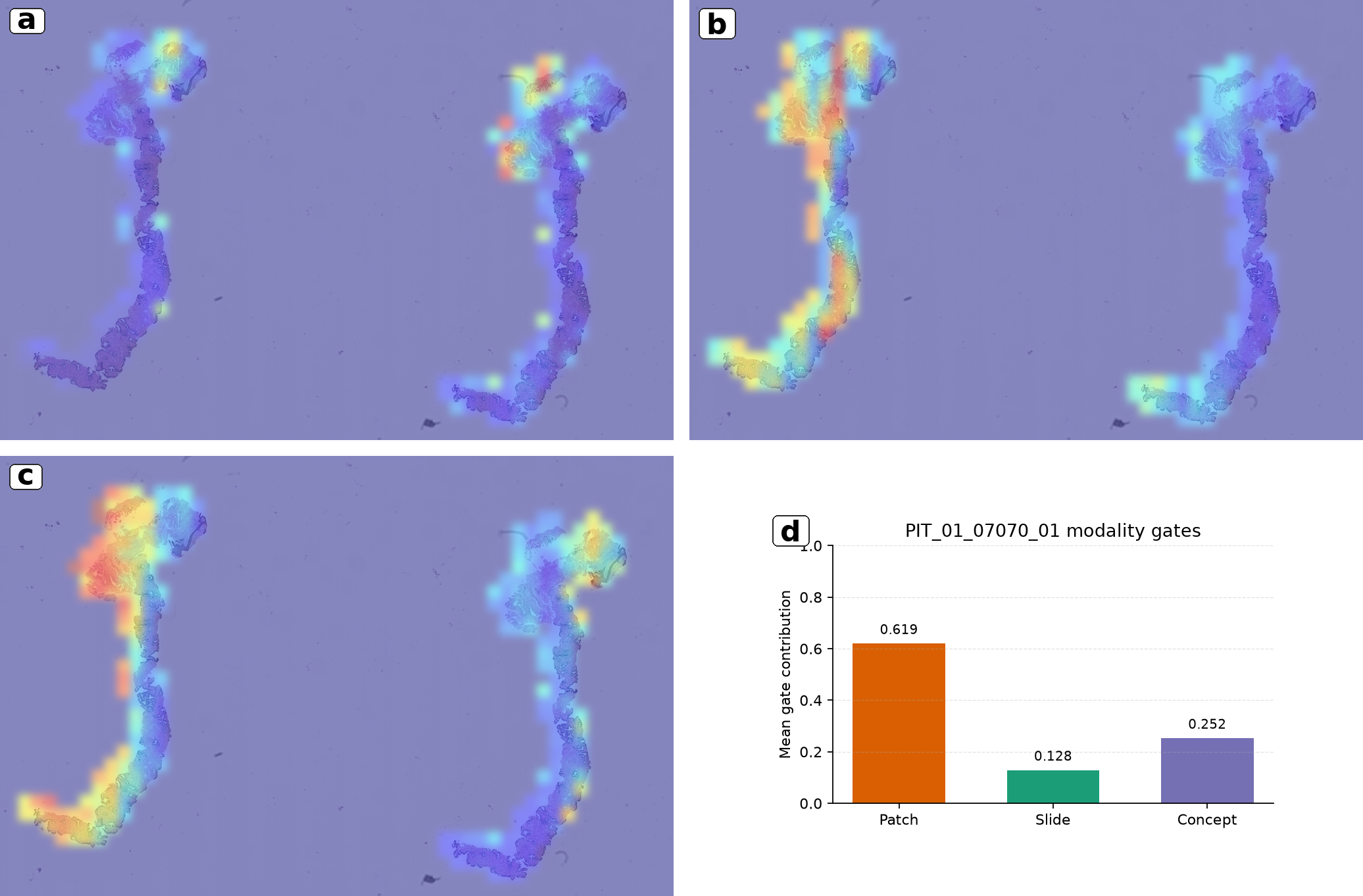}
    \caption{\textbf{Additional case-level modality analysis for case PIT\_01\_07070\_01.}
    \textbf{a--c}, Patch-, slide-, and concept-level attention maps, respectively.
    \textbf{d}, Mean decoder gate contributions for the same case. The panels
    illustrate the distinct spatial emphasis of the three representation streams
    and their relative use during report generation.}
    \label{fig:attention_gate_case_07070}
\end{figure*}

\begin{figure*}[p]
    \centering
    \includegraphics[width=\linewidth]{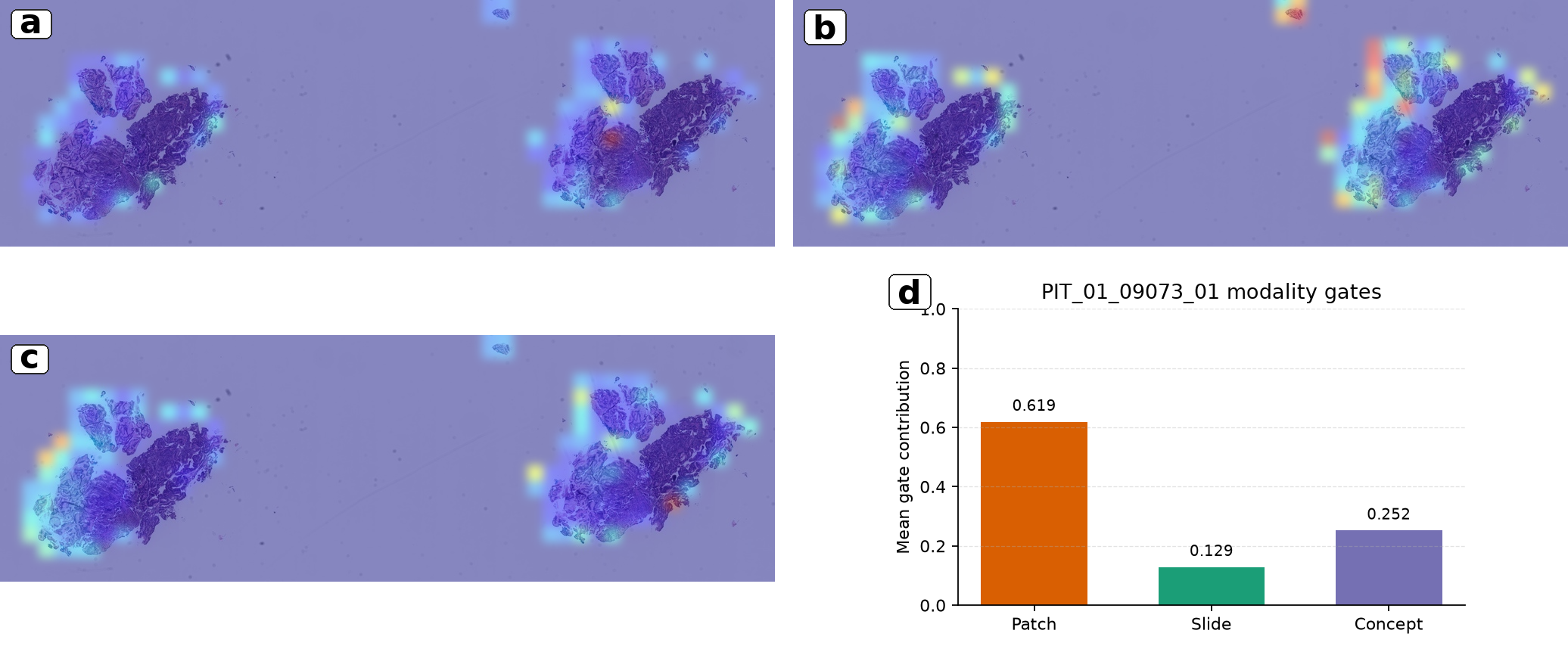}
    \caption{\textbf{Additional case-level modality analysis for case PIT\_01\_09073\_01.}
    \textbf{a--c}, Patch-, slide-, and concept-level attention maps, respectively.
    \textbf{d}, Mean decoder gate contributions for the same case. The panels
    illustrate the distinct spatial emphasis of the three representation streams
    and their relative use during report generation.}
    \label{fig:attention_gate_case_09073}
\end{figure*}

To complement the representative example in Fig.~\ref{fig:attention_gate_case},
Figs.~\ref{fig:attention_gate_case_03006}--\ref{fig:attention_gate_case_09073}
show modality-specific attention and mean decoder gate contributions for three
additional REG 2025 test cases. Each example uses the same panel organization:
\textbf{a}, patch-level attention; \textbf{b}, slide-level attention;
\textbf{c}, concept-level attention; and \textbf{d}, mean decoder gate
contributions. Across the examples, the three attention streams emphasize
different spatial regions, while all modalities retain non-zero gate
contributions.

%%%%%%%%%%%%%%%%%%%%%%%%%%%%%%%%%%%%
\subsection{Feature extraction details}
\label{app:feature_extraction_details}

\begin{figure}[t]
\centering
\includegraphics[width=0.98\textwidth]{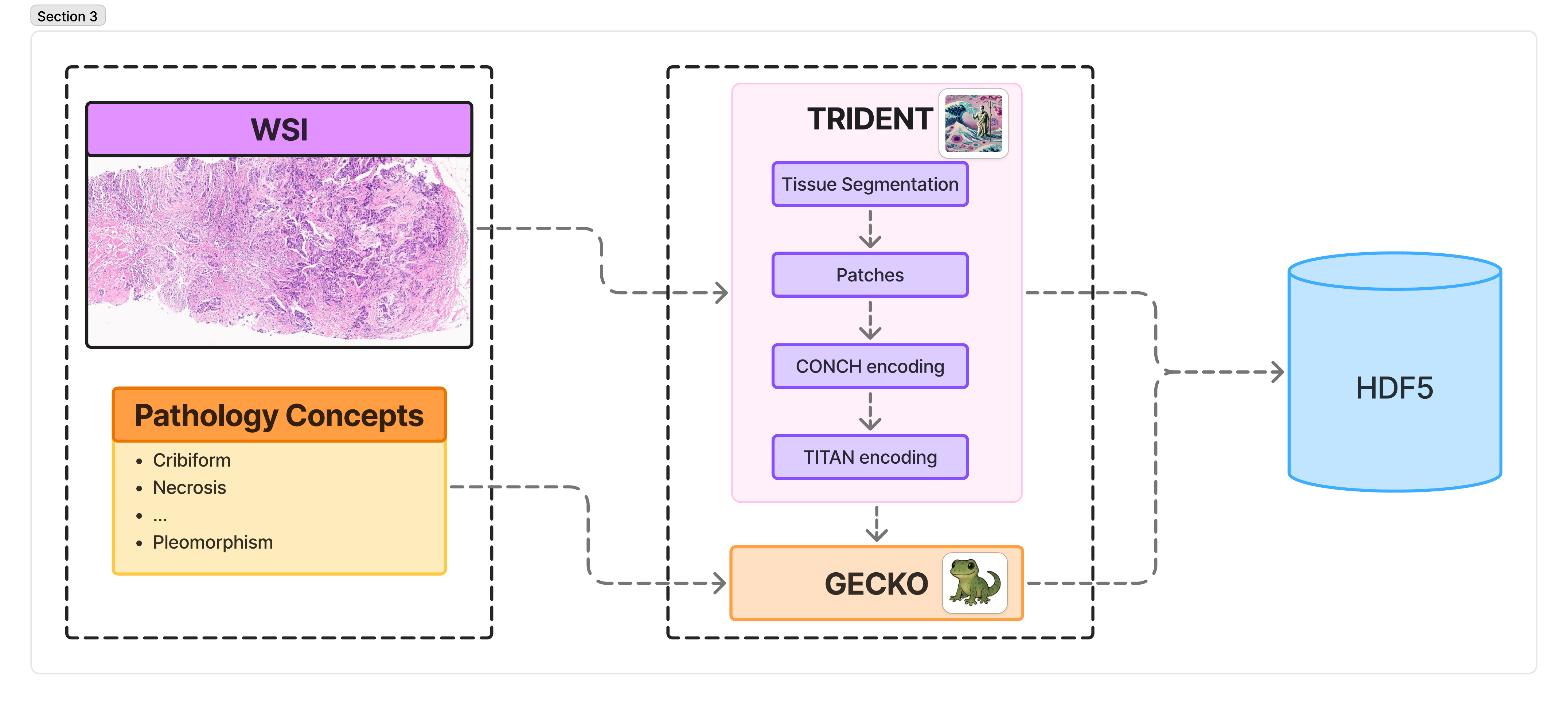}
\caption{\textbf{Offline feature extraction pipeline.} Whole-slide images are first segmented and tiled into tissue patches. Patch-level embeddings are extracted using CONCHv1.5, slide-level embeddings are obtained by aggregating patch features with TITAN, and concept-level representations are generated using GECKO from curated diagnostic concepts. All three feature streams are precomputed and stored before SCOUT training.}
\label{fig:feature_extraction_details}
\end{figure}

Figure~\ref{fig:feature_extraction_details} summarizes the offline feature extraction pipeline used to generate the patch-, slide-, and concept-level inputs for SCOUT.

We extracted all WSI-derived representations with a fixed preprocessing and encoding pipeline to ensure that differences across model components reflect the proposed fusion strategy rather than changes in the underlying visual features. Tissue regions were first identified using the TRIDENT preprocessing pipeline, which performs tissue-versus-background segmentation before downstream patching and feature extraction. TRIDENT provides a unified interface for both patch-level and slide-level pathology foundation models, including CONCHv1.5 and TITAN, and is designed to avoid unnecessary computation on non-tissue background regions \cite{trident}. After segmentation, each WSI was tiled into non-overlapping $512 \times 512$ pixel tissue patches at $20\times$ magnification. For a slide containing $N$ retained tissue patches, the patch feature matrix is therefore denoted as $X \in \mathbb{R}^{N \times 768}$.

Patch-level visual features were extracted with CONCHv1.5. CONCH is a vision--language foundation model for computational pathology that combines an image encoder, a text encoder, and a multimodal decoder, and is pretrained with contrastive image--text alignment together with image-conditioned captioning objectives \cite{conch}. This training design encourages histology image embeddings to occupy a representation space aligned with pathology language, making CONCH features suitable for downstream vision--language report generation. In our pipeline, the frozen CONCHv1.5 image encoder maps each $512 \times 512$ tissue patch to a 768-dimensional embedding, yielding the local patch representation used by SCOUT.

Slide-level context features were extracted with TITAN. TITAN is a multimodal whole-slide foundation model that uses CONCHv1.5 patch features as input and aggregates them with a ViT-based slide encoder to produce a WSI-level embedding \cite{titan}. Its pretraining combines slide-level representation learning with vision--language alignment to captions and pathology reports, allowing the resulting slide embedding to summarize global tissue architecture and diagnostic context. We use the frozen TITAN slide encoder to transform the set of CONCHv1.5 patch embeddings for each WSI into a single global slide representation.

Concept-level features were extracted with GECKO. GECKO introduces gigapixel vision--concept contrastive pretraining in which an interpretable concept prior is generated from task-relevant pathology concepts and aligned with WSI-level visual representations \cite{gecko}. Following this paradigm, we used an LLM-assisted curation process to define diagnostic concepts for the relevant cancer types, with concepts reviewed and refined to emphasize visually discriminative morphology. These concepts were used to train the GECKO concept branch, after which the trained model generated WSI-level GECKO embeddings. In SCOUT, these embeddings provide explicit semantic anchors that complement the local CONCH patch features and the global TITAN slide representation.

\subsection{Training details, computational overhead, and hyperparameters}\label{app:training_reproducibility}

For reproducibility, fixed training, validation, and test partitions were created for each dataset using an approximately 80/10/10 case-level split. These partitions were generated once and reused for every SCOUT and baseline experiment, ensuring that performance differences were not attributable to split variability. The exact split files used in all experiments are released in the project repository. Patch-level CONCHv1.5 features, slide-level TITAN features, and concept-level GECKO features were precomputed before model training and stored in HDF5 format. During training, the report-generation model therefore loads fixed feature tensors from disk rather than recomputing tissue segmentation, patch extraction, foundation-model embeddings, or concept encodings online.

This design substantially reduces computational overhead during model optimization. The expensive WSI preprocessing and foundation-model inference stages are performed once per slide as an offline preprocessing step. Training SCOUT only incurs the cost of reading the HDF5 feature files, projecting the three feature streams, applying the context-aware encoder, and decoding the target report. Consequently, the additional runtime overhead of the proposed model relative to single-feature baselines comes primarily from the lightweight slide- and concept-conditioning modules and the gated multimodal fusion mechanism, rather than from repeated feature extraction.

All training and evaluation runs were implemented using the PyTorch Lightning framework. Experiments were conducted on four NVIDIA H200 GPUs, with 32~GB memory available per GPU in our training jobs. Validation data were used for checkpoint selection and hyperparameter tuning, while the held-out test split was used only for final evaluation. Table~\ref{tab:hyperparameters} summarizes the main hyperparameters used in our experiments. The full set of hyperparameter values, configuration files, and run settings is provided in the project repository: \url{https://github.com/surykntsingh/SCOUT}.

\begin{table}[t]
\centering
\caption{\textbf{Summary of training hyperparameters.} The table reports the main optimization, sequence-length, and model-capacity settings used for SCOUT training.}
\label{tab:hyperparameters}
\begin{tabular}{l|l}
\toprule
\textbf{Hyperparameter} & \textbf{Value} \\
\midrule
Maximum sequence length & REG 2025: 320; TCGA-BRCA: 600; HistAI: 800 \\
Optimizer & AdamW \\
Learning rate & $7 \times 10^{-5}$ \\
Weight decay & $1 \times 10^{-3}$ \\
Batch size & 1 \\
Maximum epochs & 100 \\
Early stopping patience & 5 epochs \\
Learning-rate scheduler & ReduceLROnPlateau \\
Checkpoint selection criterion & Validation loss \\
Encoder/decoder hidden dimension & 768 \\
Projected feature dimension & 768 \\
Number of Transformer layers ($L$) & 4 \\
Number of attention heads & 4 \\
Dropout & 0.2 \\
\bottomrule
\end{tabular}
\end{table}

\subsection{Curated GECKO diagnostic concepts for TCGA-BRCA}\label{app:brca_gecko_concepts}

Table~\ref{tab:brca_gecko_concepts} lists the curated diagnostic concepts used to construct the GECKO concept prior for TCGA-BRCA. These concepts capture breast cancer morphology at multiple levels, including architectural growth patterns, cytologic features, stromal and inflammatory context, histologic grade, and clinically relevant subtype-associated appearances.

\refstepcounter{table}\label{tab:brca_gecko_concepts}
\noindent\textbf{Table~\thetable: Curated diagnostic concepts used for GECKO concept encoding in TCGA-BRCA.} The table lists breast cancer morphology concepts used to construct the GECKO concept prior.
\vspace{0.5em}

\begin{longtable}{>{\raggedright\arraybackslash}p{0.28\textwidth}>{\raggedright\arraybackslash}p{0.66\textwidth}}
\hline
\textbf{Concept} & \textbf{Description} \\
\hline
\endfirsthead
\hline
\textbf{Concept} & \textbf{Description} \\
\hline
\endhead
\endfoot
\hline
\endlastfoot
Ductal architecture & Tumor cells arranged in duct-like or glandular structures resembling normal breast ducts, typical of invasive ductal carcinoma. \\
Lobular architecture & Small uniform tumor cells infiltrating in single-file or targetoid patterns, characteristic of invasive lobular carcinoma. \\
Cribriform pattern & Sieve-like tumor glands with punched-out lumina separated by thin epithelial bridges. \\
Tubule formation & Well-formed tubular glands indicating glandular differentiation and lower histologic grade. \\
Solid nests or sheets & Cohesive sheets or nests of malignant cells lacking glandular lumina, typical of high-grade carcinoma. \\
Micropapillary structures & Small papillary clusters without fibrovascular cores floating in clear spaces, often linked to lymphovascular invasion. \\
True papillary structures & Fibrovascular cores lined by tumor epithelium forming delicate papillae, seen in papillary carcinoma. \\
Mucinous component & Extracellular mucin pools containing floating tumor cells, diagnostic of mucinous carcinoma. \\
Signet-ring cells & Tumor cells with intracytoplasmic mucin pushing the nucleus to the periphery. \\
Apocrine differentiation & Cells with abundant eosinophilic granular cytoplasm and prominent nucleoli resembling apocrine glands. \\
Medullary-like features & Sheets of large pleomorphic cells with dense lymphoid stroma and well-defined borders. \\
Metaplastic differentiation & Areas showing squamous, spindle cell, or cartilaginous differentiation within carcinoma. \\
Comedo necrosis & Central necrosis with granular eosinophilic debris and ghost cells within ducts, common in high-grade DCIS. \\
Tumor necrosis & Foci of coagulative cell death within invasive tumor nests, appearing as eosinophilic debris. \\
Calcifications & Basophilic granular or laminated calcium deposits within ducts or necrotic regions. \\
Nuclear pleomorphism & Marked variation in nuclear size, shape, and chromatin texture indicating high-grade malignancy. \\
Prominent nucleoli & Large, distinct nucleoli within tumor nuclei, suggesting increased transcriptional activity. \\
High mitotic activity & Frequent mitotic figures with condensed chromatin, reflecting rapid proliferation. \\
Tumor borders & Infiltrative margins with irregular extension into adipose tissue or smooth pushing borders in medullary-like tumors. \\
Adipose tissue invasion & Tumor cells invading breast fat lobules, confirming stromal invasion. \\
Desmoplastic stroma & Reactive fibroblast-rich collagenous stroma surrounding invasive tumor glands. \\
Reactive fibroblasts & Spindle-shaped fibroblasts with elongated nuclei indicating active stromal response. \\
Stromal fibrosis & Dense fibrotic stroma surrounding tumor nests, often reflecting chronic reaction. \\
Stromal edema & Loose myxoid stroma with separated collagen bundles and increased interstitial fluid. \\
Lymphocytic infiltration & Clusters of lymphocytes infiltrating around or within tumor nests, representing immune response. \\
Plasma cell infiltration & Presence of plasma cells in the stroma, often accompanying chronic inflammatory response. \\
Foamy macrophages & Lipid-laden macrophages in necrotic or regressive areas of the tumor. \\
Granulomatous reaction & Collections of epithelioid histiocytes or multinucleated giant cells near necrotic or regressed tumor tissue. \\
Hemorrhage & Extravasated red blood cells within tumor stroma or necrotic zones. \\
Lymphovascular invasion & Tumor emboli visible within endothelial-lined lymphatic or vascular channels. \\
Perineural invasion & Tumor cells encasing or infiltrating nerve bundles in breast parenchyma. \\
Basement membrane disruption & Loss or fragmentation of the myoepithelial layer and basement membrane, indicating stromal invasion. \\
Myoepithelial layer preservation & Intact peripheral myoepithelial cells around ducts, supporting in-situ lesion rather than invasion. \\
Ductal carcinoma in situ (DCIS) & Neoplastic epithelial proliferation confined within ducts with intact myoepithelial boundaries. \\
Lobular carcinoma in situ (LCIS) & Monomorphic small cells filling lobules without stromal invasion or desmoplasia. \\
Paget disease of nipple & Large malignant epithelial cells scattered within epidermis of nipple skin. \\
Inflammatory necrosis & Necrosis accompanied by mixed inflammatory infiltrate, distinct from comedo-type necrosis. \\
Tumor-infiltrating lymphocytes (TILs) & Dense immune cell aggregates within tumor nests or stroma indicating immune engagement. \\
Apoptotic bodies & Small condensed nuclear fragments scattered among viable tumor cells. \\
Atrophic ducts or glands & Small compressed benign ducts adjacent to invasive tumor foci. \\
Fibrocystic change & Dilated benign cysts and stromal fibrosis in background parenchyma. \\
Sclerosing adenosis & Proliferation of small ductules within dense fibrous stroma, sometimes mimicking invasion. \\
Radial scar & Central fibroelastic core with radiating ducts and glands creating stellate appearance. \\
Fat necrosis & Foamy macrophages and multinucleated giant cells surrounding necrotic adipocytes. \\
Vascular proliferation & Increased small blood vessels or capillaries within tumor stroma. \\
Psammoma-like calcifications & Round, laminated calcific concretions sometimes seen in papillary lesions. \\
Dermal invasion & Tumor infiltration into dermal collagen beneath epidermis. \\
Ulceration & Loss of epidermal surface with exposed tumor, often in skin-invading lesions. \\
Desmoplastic response & Reactive dense collagen deposition surrounding invasive carcinoma. \\
\hline
\end{longtable}

Table~\ref{tab:reg_histai_concepts} summarizes the broader patch-level concept prompts used for the REG 2025 and HistAI experiments. In contrast to the TCGA-BRCA-specific list above, these prompts span multiple cancer types and provide general morphology anchors for heterogeneous report-generation settings.

\refstepcounter{table}\label{tab:reg_histai_concepts}
\noindent\textbf{Table~\thetable: Broader patch-level diagnostic concept prompts used for REG 2025 and HistAI.} The table lists cancer-type-specific morphology prompts and descriptions used to construct concept-level semantic inputs for heterogeneous datasets.
\vspace{0.5em}

\begin{longtable}{>{\raggedright\arraybackslash}p{0.17\textwidth}>{\raggedright\arraybackslash}p{0.24\textwidth}>{\raggedright\arraybackslash}p{0.50\textwidth}}
\hline
\textbf{Cancer type} & \textbf{Concept} & \textbf{Description} \\
\hline
\endfirsthead
\hline
\textbf{Cancer type} & \textbf{Concept} & \textbf{Description} \\
\hline
\endhead
\endfoot
\hline
\endlastfoot
Breast & Cribriform pattern & Tumor glands forming rounded or sieve-like punched-out spaces (cribriform architecture) \\
 & Papillary structures & Epithelial fronds or projections supported by fibrovascular cores (papillary architecture) \\
 & Micropapillary architecture & Small clusters of tumor cells without fibrovascular cores, often surrounded by clear lacunar spaces \\
 & Tubular structures & Well-formed small tubules or ductules lined by tumor cells \\
 & Single-file infiltration & Discohesive tumor cells arranged in single-file linear cords (characteristic of invasive lobular carcinoma) \\
 & Desmoplastic stroma & Dense fibrous connective tissue reaction around invasive tumor nests \\
 & Nuclear pleomorphism & Variation in tumor cell nuclear size, shape, and chromatin staining \\
 & Comedonecrosis & Central necrotic debris filling ducts (high-grade DCIS with comedo-type necrosis) \\
 & Tumor-infiltrating lymphocytes & Dense lymphocytic infiltrates within tumor stroma (notably in medullary subtype) \\
Prostate & Cribriform glands & Tumor glands forming large rounded sheets with multiple perforated or punched-out lumina \\
 & Glomeruloid structures & Rounded epithelial tufts or buds protruding into glandular lumina, resembling small renal glomeruli \\
 & Fused glands & Malignant glands merging together with incomplete or back-to-back lumina (Gleason pattern 4) \\
 & Poorly formed glands & Small, irregular glandular spaces with scant luminal formation (Gleason pattern 4) \\
 & Solid sheets & Sheets of tumor cells lacking any glandular differentiation (Gleason pattern 5) \\
 & Single cells/cords & Individual tumor cells and linear cords infiltrating stroma (Gleason pattern 5) \\
 & Nuclear enlargement \& prominent nucleoli & Tumor cell nuclei are enlarged with one or more prominent nucleoli \\
 & Lymphovascular invasion & Tumor cells seen within endothelial-lined blood or lymphatic channels \\
 & Perineural invasion & Tumor cells encasing or infiltrating along nerves (a hallmark of prostate carcinoma) \\
Colorectal & Tubular/villous glands & Malignant glandular structures forming long tubular or villous projections (intestinal-type adenocarcinoma) \\
 & Papillary architecture & Tumor cells lining fibrovascular cores (often seen in villous adenomas) \\
 & Mucinous pools & Abundant extracellular mucin pooling in the stroma with floating tumor cells (mucinous carcinoma) \\
 & Signet ring cells & Tumor cells with large intracytoplasmic mucin vacuoles that push the nucleus to the cell periphery (signet-ring appearance) \\
 & Micropapillary clusters & Tight clusters of tumor cells within empty lacunae lacking fibrovascular cores (invasive micropapillary component) \\
 & Tumor budding & Isolated single tumor cells or small clusters ($\leq$4 cells) at the invasive tumor front \\
 & Dirty necrosis & Neutrophil-rich necrotic debris and cell detritus filling glandular lumina (dirty necrosis) \\
 & Desmoplastic stroma & Reactive fibrous stromal reaction around infiltrative tumor glands \\
 & Tumor-infiltrating lymphocytes & Prominent intraepithelial and stromal lymphocytic infiltrates within tumor (especially in medullary-type) \\
 & Lymphovascular invasion & Tumor cells present within lymphatic or blood vessel channels (poor prognostic feature) \\
Gastric & Glandular (intestinal) patterns & Malignant glands or tubules resembling intestinal epithelium (tubular or papillary adenocarcinoma) \\
 & Signet ring cells & Tumor cells with intracytoplasmic mucin vacuoles displacing the nucleus (diffuse-type carcinoma) \\
 & Diffuse infiltration (linitis plastica) & Diffuse infiltration of gastric wall by tumor cells, thickening and stiffening the stomach ('leather-bottle' stomach) \\
 & Mucinous pools & Pools of extracellular mucin containing tumor cells (mucinous adenocarcinoma of stomach) \\
 & Micropapillary architecture & Clusters of tumor cells in clear spaces without stromal cores (gastric micropapillary variant) \\
 & Poorly differentiated nests & Solid nests or sheets of undifferentiated tumor cells \\
 & Tumor-infiltrating lymphocytes & Dense lymphocytic infiltrates within the tumor (often seen in EBV-associated gastric cancer) \\
 & Lymphovascular invasion & Tumor cells within blood or lymphatic vessels in the gastric wall (invasion of vascular channels) \\
Lung & Lepidic growth & Tumor cells proliferating along intact alveolar walls (lepidic-predominant pattern) \\
 & Papillary structures & Tumor cells forming papillary projections with fibrovascular cores \\
 & Acinar (glandular) structures & Malignant gland/tubule formation resembling pulmonary alveoli \\
 & Solid sheets & Large solid nests or sheets of tumor cells lacking glandular or squamous features \\
 & Micropapillary clusters & Tight clusters of tumor cells floating in clear spaces without true fibrovascular cores (micropapillary adenocarcinoma) \\
 & Rosettes/organoid nests & Neuroendocrine tumor cells arranged in organoid nests or rosette-like patterns \\
 & Keratin pearls & Concentric laminated keratinous material within nests of squamous carcinoma cells \\
 & Intercellular bridges & Strands of desmosomal connections between squamous tumor cells \\
 & Necrosis (geographic) & Large zones of coagulative necrosis spanning groups of tumor cells \\
 & Tumor-infiltrating lymphocytes & Lymphocytes infiltrating tumor stroma (often seen in poorly differentiated or viral-associated lung tumors) \\
Cervical Cancer & Squamous nests & Invasive tumor islands composed of atypical squamous cells infiltrating cervical stroma \\
 & Keratin pearls & Concentric laminated keratin material within nests of squamous carcinoma cells \\
 & Koilocytosis & Perinuclear halos and nuclear atypia in squamous cells (indicative of HPV infection) \\
 & High-grade squamous dysplasia & Marked nuclear atypia and architectural disarray in surface epithelium (CIN 3) \\
 & Glandular crowding & Pseudostratified, elongated glandular epithelium with overlapping nuclei (seen in AIS) \\
 & Atypical mitoses & Bizarre or tripolar mitotic figures in dysplastic epithelium or invasive carcinoma \\
 & Stromal desmoplasia & Reactive fibrous stromal response around invasive nests \\
 & Tumor-infiltrating lymphocytes & Dense lymphocytic infiltration in stroma near tumor \\
 & Lymphovascular invasion & Tumor cells within vascular or lymphatic channels in the cervical stroma \\
 & Surface ulceration & Loss of epithelial surface with necrosis and inflammatory exudate (common in invasive SCC) \\
Bladder Cancer & Papillary architecture & Fibrovascular cores lined by multilayered atypical urothelium (papillary urothelial carcinoma) \\
 & Umbrella cells & Large superficial urothelial cells with abundant cytoplasm and oval nuclei (seen in low-grade lesions) \\
 & Thickened urothelium & Increased urothelial layers with nuclear enlargement and polarity loss (carcinoma in situ) \\
 & Discohesive tumor cells & Scattered, loosely cohesive cells with pleomorphic nuclei and mitoses \\
 & Invasion into muscularis propria & Tumor nests or cords infiltrating thick bundles of smooth muscle (muscle-invasive carcinoma) \\
 & Nested architecture & Inconspicuous bland-looking nests infiltrating deep lamina propria (nested variant) \\
 & Plasmacytoid cells & Discohesive tumor cells with eccentric nuclei and eosinophilic cytoplasm (plasmacytoid variant) \\
 & Squamous differentiation & Presence of keratinization or intercellular bridges in urothelial carcinoma \\
 & Necrosis in papillary cores & Areas of central necrosis within papillary fronds (seen in high-grade tumors) \\
 & Lymphovascular invasion & Tumor cells identified in vascular or lymphatic channels \\
\hline
\end{longtable}

%%%%%%%%%%%%%%%%%%%%%%%%%%%%%%%%%%%%%%%%%%%%%%%%%%%%%%%%%%%%

\end{document}